\documentclass{article}

\usepackage{microtype}
\usepackage{graphicx}
\usepackage{subcaption}
\usepackage{booktabs} 

\usepackage{hyperref}

\usepackage[preprint]{icml2026}

\usepackage{amsmath}
\usepackage{amssymb}
\usepackage{mathtools}
\usepackage{amsthm}

\usepackage[capitalize,noabbrev]{cleveref}

\usepackage{soul}
\usepackage{graphicx}
\usepackage{mdframed}
\usepackage{booktabs}
\usepackage{multirow}
\usepackage{wrapfig}
\usepackage{subcaption}
\usepackage{tabularx}
\usepackage{enumitem}
\usepackage{xcolor}
\usepackage{placeins}
\usepackage{titlesec}
\titlespacing*{\paragraph}{0pt}{0.3ex}{0.6em}

\theoremstyle{plain}

\theoremstyle{definition}

\theoremstyle{remark}

\usepackage[textsize=tiny]{todonotes}

\icmltitlerunning{Parameterizing Dataset Distillation via Gaussian Splatting}

\begin{document}

\twocolumn[
  \icmltitle{Parameterizing Dataset Distillation via Gaussian Splatting}

  \begin{icmlauthorlist}
    \icmlauthor{Chenyang Jiang}{hitsz,pcl}
    \icmlauthor{Zhengcen Li}{hitsz,pcl}
    \icmlauthor{Hang Zhao}{hitsz}
    \icmlauthor{Qiben Shan}{pcl}
    \icmlauthor{Shaocong Wu}{pcl}
    \icmlauthor{Jingyong Su}{hitsz,pcl}
  \end{icmlauthorlist}

  \icmlaffiliation{hitsz}{The School of Computer Science and Technology, Harbin Institute of Technology, Shenzhen, China}
  \icmlaffiliation{pcl}{Institute of Perceptual Intelligence, Pengcheng Laboratory}

  \icmlcorrespondingauthor{Chenyang Jiang}{jiangchenyang2001@qq.com}
  \icmlcorrespondingauthor{Jingyong Su}{sujingyong@hit.edu.cn}

  \icmlkeywords{Machine Learning, ICML}

  \vskip 0.3in
]



\printAffiliationsAndNotice{}  

\begin{abstract}
	Dataset distillation aims to compress training data while preserving training-aware knowledge, alleviating the reliance on large-scale datasets in modern model training.
	Dataset parameterization provides a more efficient storage structure for dataset distillation, reducing redundancy and accommodating richer information.
	However, existing methods either rely on complex auxiliary modules or fail to balance representational capacity and efficiency.
	In this paper, we propose GSDD, a simple, novel, and effective dataset parameterization technique for Dataset Distillation based on Gaussian Splatting.
	We adapt CUDA-based splatting operators for parallel training in batch, enabling high-quality rendering with minimal computational and memory overhead.
	Gaussian primitives can effectively capture meaningful training features, allowing a sparse yet expressive representation of individual images.
	Leveraging both high representational capacity and efficiency, GSDD substantially increases the diversity of distilled datasets under a given storage budget, thereby improving distillation performance.
	Beyond achieving competitive results on multiple standard benchmarks, GSDD also delivers significant performance gains on large-scale datasets such as ImageNet-1K and on video distillation tasks. In addition, we conduct comprehensive benchmarks to evaluate the computational efficiency, memory footprint, and cross-GPU architectural stability of GSDD. Code is available on https://github.com/j-cyoung/GSDatasetDistillation
\end{abstract}

\section{Introduction}


Guided by the principles of scaling laws~\cite{kaplan2020scalinglawsneurallanguage}, deep learning has advanced along the trajectory of larger models, larger datasets, and longer training, which has significantly pushed the performance boundaries of modern models. However, this paradigm also incurs enormous demands on computation, storage, and communication resources, creating a barrier that hinders both broad adoption and sustainable development. Consequently, a pivotal question in the field is how to achieve strong performance under limited computational and data resources.
%

Against this backdrop, dataset distillation has emerged as a promising solution.
This approach aims to distill the essential knowledge of large-scale datasets within a compact set of synthetic samples, or distilled images, thereby enabling models trained on the distilled dataset to achieve performance comparable to those trained on the full dataset.
Methodologically, dataset distillation formulates these synthetic images as optimizable targets, where the primary objective is to minimize the training-aware information gap between the synthetic and original data.
By adopting dataset distillation, the costs of model training, data storage, and data transmission can be substantially reduced, and can be applied in a wide range of fields, such as efficient data replay for continual learning~\cite{yang2023an,guSummarizingStreamData2024}, critical data transfer in federated learning~\cite{huang2024overcoming}, and privacy preservation~\cite{dongPrivacyFreeHow2022}.

%

Broadly, dataset distillation can be decomposed into two complementary modules: distillation algorithms and data parameterization. The former defines how training-aware key information is extracted, such as training gradients~\cite{zhaoDatasetCondensationGradient2021}, model parameters~\cite{Cazenavette_2022_CVPR}, or distributional statistics~\cite{zhaoDatasetCondensationDistribution2023}, while the latter determines the storage structure of such extracted training-aware information.
Conventional dataset distillation methods typically rely on separate raw images to construct synthetic sets, where the resulting dense pixel-wise representation leads to significant informational redundancy and limit the performance of distilled dataset under limited storage~\cite{kimDatasetCondensationEfficient2022,dengRememberDistillingDatasets2022}.

Pioneering studies have explored diverse parameterization strategies to mitigate informational redundancy by targeting either inter-image correlations or intra-image structures. 
Approaches focusing on inter-image correlations typically design specialized network architectures~\cite{dengRememberDistillingDatasets2022,liuDatasetDistillationFactorization2022,weiSparseParameterizationEpitomic2023,liuMGDDMetaGenerator2023} to encapsulate shared knowledge across the distilled set. 
However, such hand-crafted structures often limit the representation space and introduce complexity regarding both architectural design and storage estimation. 
Conversely, targeting intra-image structures represents a promising alternative~\cite{kimDatasetCondensationEfficient2022,shinFrequencyDomainBasedDataset2023,shin2025distilling}, yet existing methods in this category frequently struggle to achieve a satisfactory balance between representational capacity and computational efficiency.

In this paper, we propose a novel approach GSDD by parameterizing Distilled Dataset via Gaussian Splatting.
We find that 2D Gaussian splatting serves as an efficient representation for distilled datasets, offering strong representational capacity and computational efficiency, while maintaining structural simplicity, introducing few additional architecture design or hyperparameters.
GSDD represents each distilled image with a sparse set of 2D Gaussians. Each Gaussian explicitly encodes training-aware image features that span multiple pixels and dataset-level images could be decoded in parallel efficiently by an adaptation of rasterization cuda operators. Additionally, we quantify the importance of diversity of distilled dataset which has benn widely discussed in parameterization but lack of an official justification, and we further show how GSDD could benefit from efficiency to improve the diversity and image quality.

Our contributions can be summarized as follows:

\begin{itemize}
    \item We introduce GSDD, a novel dataset parameterization method that represents distilled images using 2D Gaussian primitives. This is the first work to integrate sparse Gaussian modeling into the field of dataset distillation.
    
    \item We develop a comprehensive pipeline for GSDD, which encompasses a specialized parameterization scheme, tailored optimization objectives, and a batch-parallel Gaussian rasterization operator.
    
    \item We provide a quantitative investigation into the dataset diversity, and demonstrate how GSDD uniquely facilitates broader sample coverage with better fidelity.
    
    \item Extensive evaluations across multiple benchmarks, including ImageNet-1K and video recognition datasets, demonstrate that GSDD achieves competitive performance while maintaining superior efficiency in both GPU memory usage and parameterization time.
\end{itemize}

\begin{figure}[t]
	\centering
	\includegraphics[width=\linewidth]{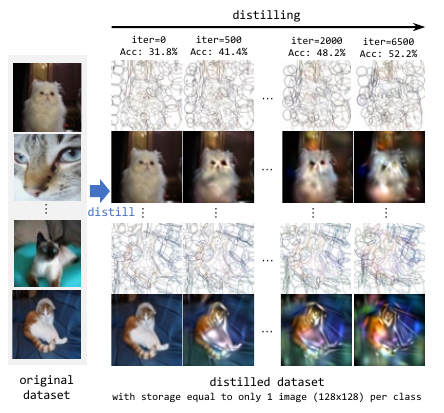}
    \caption{
    \textbf{Visualization of the GSDD distillation process.} The right columns show the evolution of the distilled dataset, with Gaussian ellipses (upper) and rendered images (lower).
	}
	\label{fig:gs_vis}
\end{figure}
\section{Background and Motivation}

\subsection{Related Work}

\paragraph{Dataset Distillation Algorithms}
Modern deep learning models typically rely on large-scale training datasets, whereas dataset distillation aims to replace the original dataset with a much smaller set of high-quality synthetic samples~\cite{wangDatasetDistillation2018} while keeping knowledge in the original dataset. 
Formally, let the original dataset be $\mathcal{T}=\{(t_i,y_i)\}_{i=1}^{N_\mathcal{T}}\subseteq\mathcal{D}$, where $t_i$ and $y_i$ denote the input sample and its corresponding label, and $\mathcal{D}$ represents the underlying data distribution. 
Without loss of generality, we omit labels in the following and simply denote the synthetic dataset as $\mathcal{S}=\{s_j\}_{j=1}^{N_\mathcal{S}}$.
Dataset distillation seeks to construct a compact synthetic dataset $\mathcal{S}=\{(s_j,y_j)\}_{j=1}^{N_\mathcal{S}}$, such that models trained on $\mathcal{S}$ can achieve comparable performance to those trained on $\mathcal{T}$:
 {
	\begin{gather}
	\mathcal{S}=\arg\min_{\mathcal{S}}\ \big|l(\theta_\mathcal{S},\mathcal{D})-l(\theta_\mathcal{T},\mathcal{D})\big|,\notag\\
	\theta_\mathcal{S}=\arg\min_{\theta}l(\theta,\mathcal{S}), \quad
	\theta_\mathcal{T}=\arg\min_{\theta}l(\theta,\mathcal{T}).\notag
	\end{gather}
 }


In dataset distillation, synthetic data are treated as learnable objects, where the pixel values of distilled images are directly optimized as parameters. The introduction of the different algorithms is detailed in Appendix~\ref{related_works}. 

\paragraph{Parameterization of Dataset Distillation}

Beyond pixel-level representations, a number of parameterization techniques have been proposed to improve storage efficiency, increase data diversity, and further enhance distillation performance, which can be broadly categorized into intra-image and inter-image redundancy reduction approaches. 
\textbf{The former} commonly represent images using combinations of shared bases or latent codes together with decoders, aiming to factorize information shared across images while preserving image-specific details~\cite{liuDatasetDistillationFactorization2022,dengRememberDistillingDatasets2022,liuMGDDMetaGenerator2023,weiSparseParameterizationEpitomic2023}.
However, the integration of such networks may introduce additional memory and computational overhead~\cite{cazenavetteGeneralizingDatasetDistillation2023}. 
More importantly, these designs can complicate storage estimation and hyperparameter selection, which might, in some cases, limit representation flexibility. 
For instance, the storage footprint of SPEED~\cite{weiSparseParameterizationEpitomic2023} is determined by a complex interplay of hyperparameters: 
$DK + 1.5NHk + R(3D^2 + 7D) + L(D + 1)$. 
Under a fixed storage budget, finding an optimal configuration can be challenging, as it requires appropriate value ranges and a delicate balance between multiple parameters, such as the token dimension $D$ and the network depth $R$.
Furthermore, certain approaches may even struggle to parameterize the dataset due to network structure constraint under stringent combination of storage budgets and image resolution~\cite{liuDatasetDistillationFactorization2022}.
\textbf{The latter}~\cite{kimDatasetCondensationEfficient2022,shinFrequencyDomainBasedDataset2023,shin2025distilling} leverages differential transformations to reduce intra-image redundancy. 
For instance, FreD~\cite{shinFrequencyDomainBasedDataset2023} optimizes the coefficients of Fourier bases; while computationally efficient, its representational capacity is somewhat constrained by the use of fixed bases. 
In contrast, DDiF~\cite{shin2025distilling} employs Implicit Neural Representations (INR) to model images, achieving superior performance through an expanded optimization space, albeit at the cost of efficiency due to the nature of point-wise querying. 
Although these approaches utilize different methodologies, they share a common principle of enhancing performance by increasing diversity, which corresponds to the potential volume of representable images enabled by the parameterization scheme, as illustrated in Figure~\ref{fig:performance_diversity_cross_dataset}.



\paragraph{Gaussian Splatting} Gaussian splatting~\cite{3dgs} was proposed to represent 3D scenes in the field of novel view synthesis, offering higher rendering efficiency and explicit geometric control compared to NeRF-based representations~\cite{NeRF}. 3DGS~\cite{3dgs} initializes a set of Gaussian primitives within a 3D scene, which are then rendered from multiple viewpoints and optimized to match the corresponding ground-truth images, enabling arbitrary-view synthesis. In recent years, Gaussian splatting has been extended to a wide range of domains, including digital humans~\cite{qian20243dgs,zielonka2025drivable}, autonomous driving~\cite{zhou2024drivinggaussian,huang2024textit}, dynamic scene modeling~\cite{wu20244d,huang2024sc}, and 3D editing~\cite{chen2024gaussianeditor}. More recently, Gaussian representations have also been applied to 2D vision tasks such as image super-resolution~\cite{hu2024gaussiansrhighfidelity2d,chen2025generalizedefficient2dgaussian} and image representation~\cite{zhu2025largeimagesgaussianshighquality,weiss2024gaussianbillboardsexpressive2d,gaussianimage}.
Although Gaussian-based paradigms offer potential for image representation, their capacity to encode critical training-aware features remains unexplored. Additionally, the absence of adapted Gaussian splatting kernels capable of rendering independent images in parallel within a batch hinders the practical application in dataset distillation.


\subsection{Motivation}

\begin{figure}[t]
	\centering
	\includegraphics[width=\linewidth]{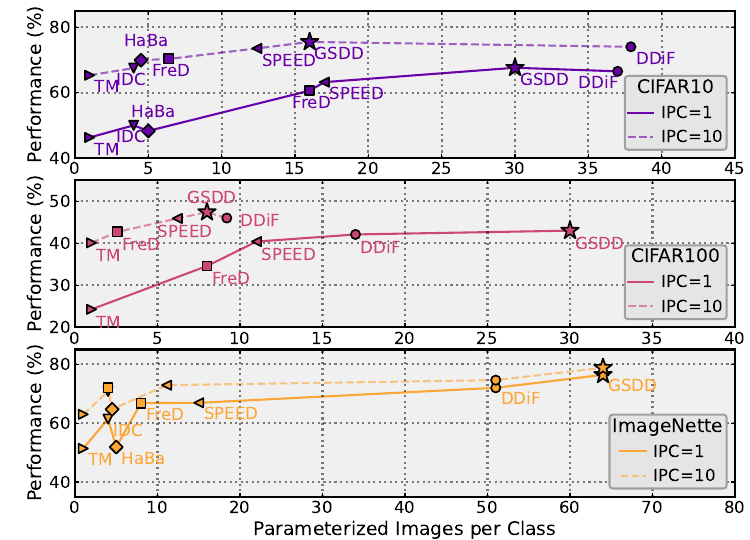}
	\caption{
	\textbf{Scaling Behavior with Parameterized Image Count.}
	Model performance scales positively with the number of parameterized images under fixed storage budget (IPC=1 or 10) in the parameterization literature, including TM~\cite{Cazenavette_2022_CVPR}, IDC~\cite{kimDatasetCondensationEfficient2022}, HaBa~\cite{liuDatasetDistillationFactorization2022}, FreD~\cite{shinFrequencyDomainBasedDataset2023}, SPEED~\cite{weiSparseParameterizationEpitomic2023}, DDiF~\cite{shin2025distilling}. 
	}
	\label{fig:performance_diversity_cross_dataset}
	\vspace{-0.3cm}
\end{figure}

\begin{figure*}[t]
	\centering
	\includegraphics[width=0.99\linewidth]{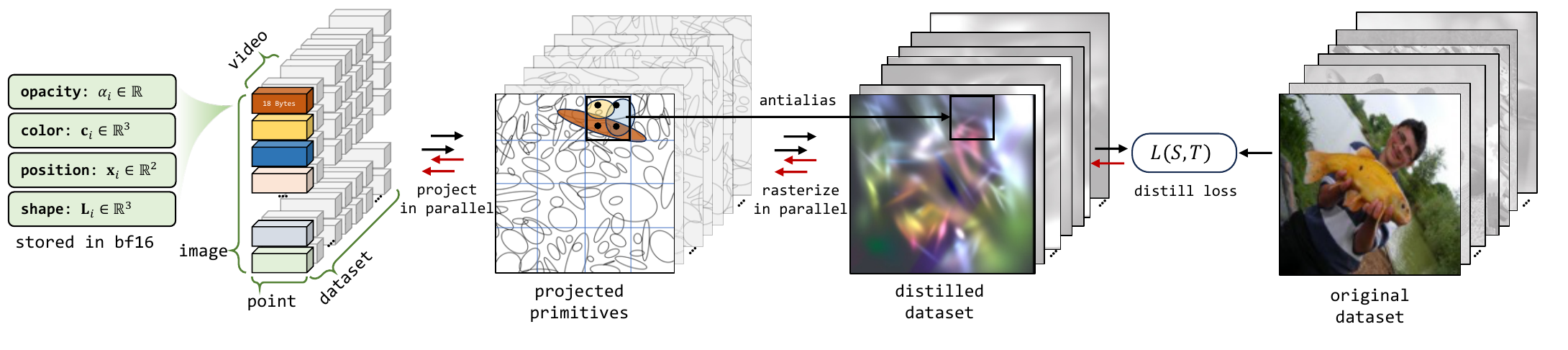}
	\caption{\small \textbf{Framework of GSDD}. Each image is parameterized by a collection of Gaussian primitives, where each primitive is defined by four attributes: position, shape, color, and opacity. Video modalities are similarly treated as temporal sequences of frames. During the optimization process, these primitives are synthesized into images via a massively parallelized differentiable rasterizer, which incorporates antialiasing to ensure high-fidelity rendering. The distillation loss is then computed between the synthesized and real images, and the gradient is finally backpropagated to the Gaussian parameters.}
	\label{fig:framework}
	\vspace{-0.2cm}
\end{figure*}

We find Gaussian representation is an efficient approach to parameterize the distilled dataset due to its structural simplicity and inherent adaptability.
In this framework, each distilled image is represented as a weighted mixture of Gaussian primitives, where the synthesis process can be formulated as the joint optimization of Gaussian \textit{bases} and their corresponding \textit{coefficients}. The design of these Gaussian primitives is straightforward, requiring no auxiliary parameterization by carefully-designed network and thus facilitating predictable storage overhead and a fine-grained control of storage. Furthermore, this explicit representation enables efficient rendering of distilled images through highly parallelized rasterizers, effectively mitigating the computational and memory overheads often incurred by sophisticated parameterization. From a representation perspective, a single Gaussian primitive can characterize local cross-pixel regions with a minimal set of parameters, thereby avoiding informational redundancy.
As illustrated in Figure~\ref{fig:gs_vis}, as the distillation iterations progress, the Gaussian points increasingly concentrate on the primary objects and structural boundaries, while the background regions are sparsely represented.
Such effectively reallocating the representational budget toward task-relevant areas within the synthetic images. 
Leveraging the parameterization efficiency, we further extend Gaussian representations to ImageNet-scale dataset distillation and video modalities.

As illustrated in Figure~\ref{fig:performance_diversity_cross_dataset}, prevailing paradigms for distilled dataset parameterization typically enhance performance by scaling the number of representable images. However, an excessive quantity of distilled images often leads to a significant degradation in individual image fidelity and imposes prohibitive computational overhead. In contrast, Gaussian representations exhibit superior representational capacity and efficiency. This allows for an increase in the volume of distilled images within a fixed storage budget without compromising representation quality or incurring excessive computational costs, thereby effectively improving the overall performance of the distilled dataset.


\section{Methodology}
\label{method}

\subsection{Parameterization of Dataset Distillation via Gaussian Splatting}

We adopt a 2D Gaussian mixture model to parameterize distilled images. Each distilled image \(s_j\) is represented as a set of \(M\) Gaussian components, denoted by  $G_j=\{g_k \mid k=1,\cdots,M\}$.
Specifically, for a pixel located at coordinates \((x,y)\), the contribution from the \(k\)-th Gaussian component \(g_k\) is defined by the following unnormalized Gaussian function:
{
\small
\begin{equation}
g_k(x,y;\mu_k,\Sigma_k)=\exp\!\left(-\tfrac{1}{2}(\mathbf{x}-\mu_k)^T \Sigma_k^{-1}(\mathbf{x}-\mu_k)\right),
\end{equation}
}
where \(\mathbf{x}=[x,y]^T\) denotes the pixel center and \(\mu_k=[u_k,v_k]^T\) denotes the learned centroid (or mean vector) of the Gaussian component. The covariance matrix \(\Sigma_k\) determines the size, shape, and orientation of the Gaussian. To ensure that \(\Sigma_k\) remains positive semi-definite during optimization, it's parameterized by Cholesky decomposition, \(\Sigma_k=L_kL_k^T\), where the lower-triangular matrix \(L_k\) is given by $[l_{11}^k,l_{21}^k,l_{22}^k]$.

Each Gaussian \(g_k\) is associated with a color vector \(\mathbf{c}_k \in \mathbb{R}^3\) and an opacity \(\alpha_k\). The final color of pixel \((x,y)\) is then synthesized by aggregating the contributions of all Gaussians:  
\begin{equation}
\mathbf{c}(x,y)=\sum_{k=1}^M \alpha_k \cdot g_k(x,y;u_k,v_k,\Sigma_k)\cdot \mathbf{c}_k.
\end{equation}

Each Gaussian primitive is characterized by four attributes: position, shape, color, and opacity, and is represented by a 9-dimensional parameter vector $p_k = (u_k, v_k, l_{11}^k, l_{21}^k, l_{22}^k, \mathbf{c}_k^r, \mathbf{c}_k^g, \mathbf{c}_k^b, \alpha_k) \in \mathbb{R}^9$.
Accordingly, the parameter set of an entire distilled image $s_j$ is given by $\mathbf{p}^j=\{p_k^j\}_{k=1}^M$, and the parameterization of the whole distilled dataset $\mathcal{S}$ is $\mathcal{P}_\mathcal{S}=\{\mathbf{p}^j\}_{j=1}^{N_\mathcal{S}}.$
The parameterization process can be defined as a synthesis function $R(\cdot;\mathcal{P_S})$, which maps coordinate grid $\mathcal{C}$ to the distilled dataset $\mathcal{S}=R(\mathcal{C};\mathcal{P_S})$.

This Gaussian-based parameterization is orthogonal to the choice of dataset distillation algorithm. Once a differentiable synthesis function is defined, it can be seamlessly integrated into any distillation framework. The optimization objective is to find the optimal distilled parameters $\mathcal{P_S}^*$ that minimize the distillation loss \(L_{\text{distill}}\):  
\begin{equation}
\mathcal{P_S}^*=\arg\min_{\mathcal{P_S}}L_{\text{distill}}(\mathcal{S},\mathcal{T}),
\end{equation}
where \(L_{\text{distill}}\) can correspond to any distillation objective.  

Figure~\ref{fig:framework} illustrates the pipeline of GSDD. 
The process begins with the projection of Gaussian primitive attributes onto the pixel space to determine essential geometric properties. 
Subsequently, the color of each pixel in the synthetic images is computed in parallel through differentiable rasterizer to generate the final distilled dataset. 
Finally, the distillation loss is evaluated between the distilled dataset and the original dataset, which facilitates the end-to-end optimization of the Gaussian parameters via backpropagation.


\begin{figure*}[t]
	\centering
	\begin{subfigure}[b]{0.33\linewidth}
		\centering
		\includegraphics[width=\linewidth, trim={0mm 0 0mm 0}, clip]{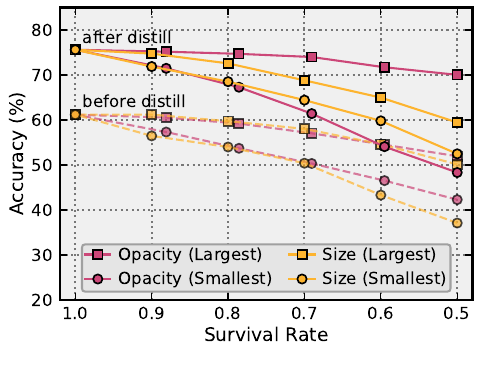}
		\caption{Effect of Gaussian pruning.}
		\label{fig:motivation_prune}
	\end{subfigure}
	\begin{subfigure}[b]{0.33\linewidth}
		\centering
		\includegraphics[width=\linewidth, trim={0mm 0 0mm 0}, clip]{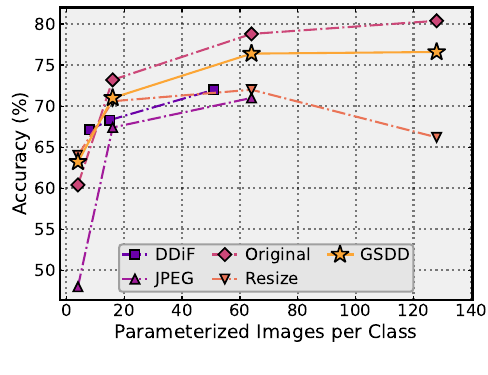}
		\caption{Scaling with distilled image diversity.}
		\label{fig:motivation_convergence}
	\end{subfigure}
	\begin{subfigure}[b]{0.33\linewidth}
		\centering
		\includegraphics[width=\linewidth, trim={0mm 0 0mm 0}, clip]{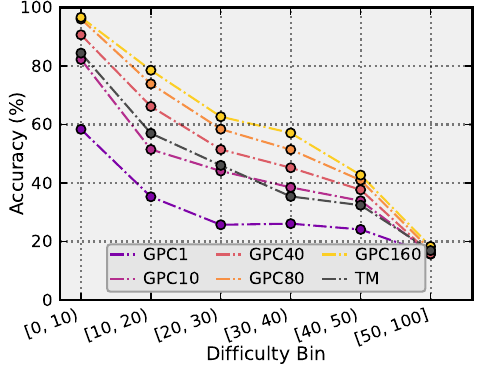}
		\caption{Accuracy across sample difficulty levels.}
		\label{fig:motivation_difficulty}
	\end{subfigure}
\caption{\small
	\textbf{Analyzing GSDD through sparsity, diversity scaling, and difficulty-aware performance.}
	\textbf{(a)} Distilled performance as a function of the remaining Gaussian primitive ratio under various pruning strategies, where ``Opacity (Largest)'' denotes the selection of primitives with the highest opacity values.
	\textbf{(b)} Test accuracy under a fixed storage budget (IPC=1) with varying numbers of parameterized images per class. The Resize baseline applies spatial resizing as in IDC, while JPEG adjusts the compression factor for different distilled dataset sizes to meet the storage constraint.
    \textbf{(c)} Accuracy on test samples categorized by difficulty levels for distilled datasets with varying image quantities under a fixed storage budget, where sample difficulty is evaluated by the model prediction error frequency.
}
	\label{fig:overall}
	\vspace{-0.2cm}
\end{figure*}

\subsection{Efficient Gaussian Splatting Design}
\label{subsec:design}

The core of the above optimization lies in the differentiable renderer $R(\mathbf{p}^j;\mathcal{C})$, which render Gaussian primitives to distilled images with both high visual fidelity and computational scalability. To enable efficient processing of Gaussian representations for large-scale dataset distillation, we incorporate several critical designs into the rendering pipeline, including parallel rendering of multiple distilled images, anti-aliasing strategies, spatial constraints on Gaussian positions and quantization.

Existing open-source rasterization operators are primarily designed for 3D scene reconstruction or single-image tasks such as representation learning and super-resolution, and generally lack support for multi-image rendering. To fully leverage the parallelism of modern GPUs, we implement customized data structures and rendering kernels for both the forward and backward passes. Specifically, all Gaussian primitives across the entire distilled dataset are represented as a single contiguous 1D vector, and a globally unique ID (GUID) system is used to assign threads that rasterize multiple images in parallel. Video distillation can also be formulated as a specific instance within this framework by representing a video sequence as a series of individual frames. Details are provided in Appendix~\ref{appendix:parallel}, and code will be attached in the supplementary material.

Some Gaussians may become highly anisotropic during optimization to capture high-frequency features such as edges, which necessitates anti-aliasing strategies. We adopt two complementary techniques: (i) analytic pre-filtering, which approximates each Gaussian’s integral over a pixel area by convolving it with a unit pixel box filter, resulting in a modified covariance:  
\begin{equation}
	\Sigma_k' = \Sigma_k + \Sigma_{box}, \quad \text{where} \quad \Sigma_{box} = \text{diag}\left(\tfrac{1}{12}, \tfrac{1}{12}\right),
\end{equation}
introducing a minimum rendering variance that suppresses aliasing for extremely narrow Gaussians; and (ii) 2$\times$2 supersampling anti-aliasing (SSAA), which averages multiple samples within each pixel area to smooth boundaries and reduce high-frequency artifacts with minimal overhead.

During optimization, some Gaussian centers tend to drift outside the normalized coordinate space \([-1,1]^2\), where they receive no gradients and become unrecoverable, a phenomenon we term \textit{Gaussian Escape}. To alleviate this, we apply a boundary regularization loss to slightly encourage Gaussian centers to remain within the viewable region:
\begin{equation}
	l_{\text{boundary}} = -\mathbb{E}_k\big[\log(1-\tilde{x}^2)+\log(1-\tilde{y}^2)\big].
	\label{eq:boundary}
\end{equation}
Visualizations of the anti-aliasing strategy and Gaussian escape are provided in Figures~\ref{fig:aliasing_cifar} and~\ref{fig:gaussian_boundary_escape} in Appendix~\ref{visualization_antialias_escape}.
Gaussian parameters exhibit robustness to reduced numerical precision, as the rasterization process does not suffer from cumulative precision errors observed in neural netowrks. Based on this, we store all Gaussian parameters in bfloat16 (bf16) precision. During training, parameters are maintained in fp32 to ensure accurate updates, while bf16 casting is applied during the forward pass to allow the model to adapt to quantization effects. After training, all parameters are quantized to bf16.

\subsection{Analysis of the Gaussian Parameterization}

\begin{table*}[t]
	\centering
	\begin{minipage}[t]{0.49\textwidth}
		\centering  \caption{\small Test accuracy on ImageNet subsets ($128\times128$) with TM. Results are averaged over 5 random seeds.}
		\label{tab:subset-mtt-ipc1}
		\resizebox{\linewidth}{!}{
			\setlength{\tabcolsep}{0.9pt}
			\renewcommand{\arraystretch}{1.3}
    \begin{tabular}{rrcccccc}
    \toprule
          & \multicolumn{1}{c}{Subset} & Nette & Woof  & Fruit & Yellow & Meow  & Squawk \\
    \midrule
          & \multicolumn{1}{c}{Original} & 87.4  & 67.0  & 63.9  & 84.4  & 66.7  & 87.5  \\
    \midrule
    \multicolumn{1}{c}{Input sized} & \multicolumn{1}{c}{TM (Vanilla)} & 51.4  & 29.7  & 28.8  & 47.5  & 33.3  & 41.0  \\
          & \multicolumn{1}{c}{FRePo} & 48.1  & 29.7  & —     & —     & —     & — \\
    \multicolumn{1}{c}{Static} & \multicolumn{1}{c}{IDC} & 61.4  & 34.5  & 38.0  & 56.5  & 39.5  & 50.2  \\
          & \multicolumn{1}{c}{FreD} & 66.8  & 38.3  & 43.7  & 63.2  & 43.2  & 57.0  \\
    \midrule
    \multicolumn{1}{c}{Parameterized} & \multicolumn{1}{c}{HaBa} & 51.9  & 32.4  & 34.7  & 50.4  & 36.9  & 41.9  \\
          & \multicolumn{1}{c}{SPEED} & 66.9  & 38.0  & 43.4  & 62.6  & 43.6  & 60.9  \\
          & \multicolumn{1}{c}{NSD} & 68.6  & 35.2  & 39.8  & 61.0  & 45.2  & 52.9  \\
    \midrule
    \multicolumn{1}{c}{Generative Prior} & \multicolumn{1}{c}{GLaD} & 38.7  & 23.4  & 23.1  & —     & 26.0  & 35.8  \\
          & \multicolumn{1}{c}{H-GLaD} & 45.4  & 28.3  & 25.6  & —     & 29.6  & 39.7  \\
    \midrule
    \multicolumn{1}{c}{Function} & \multicolumn{1}{c}{DDiF} & 72.0  & 42.9  & 48.2  & 69.0  & 47.4  & 67.0  \\
    \midrule
    \multicolumn{1}{c}{Primitive} & \multicolumn{1}{c}{GSDD} & \textbf{74.4} & \textbf{44.5} & \textbf{48.8} & \textbf{71.2} & \textbf{52.2} & \textbf{71.7} \\[-1.5ex]
          &       & \fontsize{9pt}{8pt}\selectfont±1.2  & \fontsize{9pt}{8pt}\selectfont±1.6  & \fontsize{9pt}{8pt}\selectfont±1.7  & \fontsize{9pt}{8pt}\selectfont±1.4  & \fontsize{9pt}{8pt}\selectfont±1.2  & \fontsize{9pt}{8pt}\selectfont±1.2 \\
    \bottomrule
    \end{tabular}%
		}
	\end{minipage}
	\hfill
	\begin{minipage}[t]{0.49\textwidth}
		\centering  \caption{\small Test accuracy on ImageNet subsets ($128\times128$) with gradient matching (DC) and distribution matching (DM).}
		\label{tab:subset-dm-dc-ipc1}
		\resizebox{\linewidth}{!}{
			\setlength{\tabcolsep}{2.8pt}
			\renewcommand{\arraystretch}{0.92}
			\begin{tabular}{cccccccc}
			\toprule
			\multicolumn{1}{c}{DC} & \multicolumn{6}{c}{ImageNet Subset($128\times 128$)} & \multirow{2}[2]{*}{Avg} \\
				& Nette & Woof  & Fruit & Yellow & Meow  & Squawk &  \\
			\midrule
			\multicolumn{1}{c}{DC (Vanilla) } & 34.2  & 22.5  & 21.0  & 37.1  & 22.0  & 32.0  & 28.1  \\
			\multicolumn{1}{c}{GLaD} & 35.4  & 22.3  & 20.7  & —     & 22.6  & 33.8  & 27.0  \\
			\multicolumn{1}{c}{H-GLaD} & 36.9  & 24.0  & 22.4  & —     & 24.1  & 35.3  & 28.5  \\
			\multicolumn{1}{c}{IDC} & 45.4  & 25.5  & 26.8  & —     & 25.3  & 34.6  & 31.5  \\
			\multicolumn{1}{c}{FreD} & 49.1  & 26.1  & 30.0  & —     & 28.7  & 39.7  & 34.7  \\
			\multicolumn{1}{c}{DDiF} & 61.2  & 35.2  & 37.8  & —     & 39.1  & 54.3  & 45.5  \\
			\midrule
			\multicolumn{1}{c}{GSDD} & \textbf{69.0} & \textbf{40.2} & \textbf{46.2} & \textbf{63.1}  & \textbf{40.5}  & \textbf{70.1} & \textbf{54.9} \\
				& ±0.5  & ±1.7  & ±1.6  & ±2.0  & ±1.5  & ±1.1  &  \\
			\midrule
			\multicolumn{1}{c}{DM} &       &       &       &       &       &       &  \\
			\multicolumn{1}{c}{DM (Vanilla)} & 30.4  & 20.7  & 20.4  & 36.0  & 20.1  & 26.6  & 25.7  \\
			\multicolumn{1}{c}{GLaD} & 32.2  & 21.2  & 21.8  & —     & 22.3  & 27.6  & 25.0  \\
			\multicolumn{1}{c}{H-GLaD} & 34.8  & 23.9  & 24.4  & —     & 24.2  & 29.5  & 27.4  \\
			\multicolumn{1}{c}{IDC} & 48.3  & 27.0  & 29.9  & —     & 30.5  & 38.8  & 34.9  \\
			\multicolumn{1}{c}{FreD} & 56.2  & 31.0  & 33.4  & —     & 33.3  & 42.7  & 39.3  \\
			\multicolumn{1}{c}{DDiF} & 69.2  & 42.0  & 45.3  & —     & 45.8  & 64.6  & 53.4  \\
			\midrule
			\multicolumn{1}{c}{GSDD} & \textbf{71.6} & \textbf{42.1} & \textbf{50.1} & \textbf{67.3} & \textbf{47.0} & \textbf{72.2} & \textbf{58.4} \\
				& ±0.8  & ±0.9  & ±1.0  & ±1.4  & ±0.9  & ±1.0  &  \\
			\bottomrule
			\end{tabular}%
		}
	\end{minipage}
\end{table*}

GSDD enables adaptive representation of distilled images using a small parameter set. During the distillation process, each Gaussian primitive is optimized to characterize a localized pixel region. By covering the entire image with a sparse collection of these primitives, our method effectively minimizes intra-image redundancy. In Figure~\ref{fig:overall}(a), we evaluate various pruning schemes based on different Gaussian attributes both before and after distillation. The results demonstrate that post-distillation configurations consistently outperform their pre-distillation counterparts across all pruning modes. Notably, we observe that opaque and large-scale Gaussian primitives are critical for maintaining performance. Furthermore, the role of opacity in preserving accuracy becomes more pronounced after distillation. This indicates that throughout the distillation process, specific Gaussian primitives evolve by increasing their scale and opacity, thereby effectively encoding discriminative features that are essential for downstream model training.


The diversity of distilled images, defined as the number of representable images within a fixed storage budget, exerts a direct influence on model performance. As illustrated in Figure~\ref{fig:performance_diversity_cross_dataset}, prevailing parameterization methods generally exhibit a positive correlation between dataset diversity and distilled performance. However, a fundamental trade-off exists where increasing diversity necessitates a reduction in the parameter count per image, which may constrain the expressive power of the latent representation space. We further investigate this relationship within the GSDD framework. In Figure~\ref{fig:overall}(b), we evaluate the performance of various methods across different image counts. While most methods scale positively with diversity, DDiF is constrained by its prohibitive memory and computational overhead, which limits its scalability. Traditional baselines like JPEG and Resize also show clear limitations. JPEG is hindered by compression-induced information loss, while Resize suffers from a significant degradation in representational fidelity, leading to performance decay at higher image counts. These results underscore that GSDD, leveraging its superior computational efficiency and representational capacity, converts increased diversity into performance gains.

\begin{table}[!tbp]
	\centering
	\caption{Results on Video Dataset Distillation (DM) for $112\times 112$.}
	\resizebox{\linewidth}{!}{
	  \begin{tabular}{c|cc|cc}
	  \toprule
	  Dataset & \multicolumn{2}{c|}{MiniUCF} & \multicolumn{2}{c}{HMDB51} \\
	  IPC   & 1     & 5     & 1     & 5 \\
	  \midrule
	  Full Dataset & \multicolumn{2}{c|}{62.7 ± 0.6} & \multicolumn{2}{c}{31.1 ± 0.3} \\
	  \midrule
	  Random & 9.9 ± 0.8 & 22.9 ± 1.1 & 4.6 ± 0.5 & 6.6 ± 0.7 \\
	  Herding & 12.7 ± 1.6 & 25.8 ± 0.3 & 3.8 ± 0.2 & 8.5 ± 0.4 \\
	  K-Center & 11.5 ± 0.7 & 23.0 ± 1.3 & 3.1 ± 0.1 & 5.2 ± 0.3 \\
	  \midrule
	  DM    & 15.3 ± 1.1 & 25.7 ± 0.2 & 6.1 ± 0.2 & 8.0 ± 0.2 \\
	  MTT   & 19.0 ± 0.1 & 28.4 ± 0.7 & 6.6 ± 0.5 & 8.4 ± 0.6 \\
	  FRePo & 20.3 ± 0.5 & 30.2 ± 1.7 & 7.2 ± 0.8 & 9.6 ± 0.7 \\
	  Static-DC & 13.7 ± 1.1 & 24.7 ± 0.5 & 5.1 ± 0.9 & 7.8 ± 0.4 \\
	  DM + VD & 17.5 ± 0.1 & 27.2 ± 0.4 & 6.0 ± 0.4 & 8.2 ± 0.1 \\
	  DM + GSDD & 50.8 ± 0.6 & 59.1 ± 1.3 & 19.5 ± 0.7 & 30.1 ± 1.5 \\
	  \bottomrule
	  \end{tabular}%
	}
	\label{tab:video_result}%
	\vspace{-0.5cm}
  \end{table}%

The distillation of samples across varying difficulty levels remains a central challenge in the field~\cite{guoLosslessDatasetDistillation2024,chenDataDistillationCan2024,sunDiversityRealismDistilled2024}. Increased diversity expands the coverage of medium-difficulty samples, a feat traditional methods fail to achieve with limited image budgets, thereby hindering the transition toward lossless distillation. We explore the nexus between diversity and sample coverage in Figure~\ref{fig:overall}(c), where sample difficulty is quantified by the prediction error frequency across 100 independently trained models. Our analysis reveals that while initial increases in diversity significantly bolster performance on easy samples, sustained scaling yields the most pronounced improvements for medium-difficulty samples. In contrast, extremely difficult samples, which likely consist of mislabeled or out-of-distribution (OOD) data, derive negligible benefit from increased diversity, suggesting they are inherently less learnable in the distillation process.

\section{Experiments}
\FloatBarrier
\begin{table*}[t]
	\centering
	\begin{minipage}[t]{0.405\textwidth}
		\centering\caption{Results on ImageNet-subsets with Distribution Matching (DM) for $256\times 256$.}
		\label{tab:subset-dm256-ipc1}
		\resizebox{\linewidth}{!}{
			\setlength{\tabcolsep}{1.0pt}
			\renewcommand{\arraystretch}{1.12}
			{
			\begin{tabular}{ccccccc}
			\toprule
			{Method} & Nette & Woof  & Fruit & Yellow & Meow  & Squawk \\
			\midrule
			{Vanilla} & {32.1 } & {20.0 } & {19.5 } & {33.4 } & {21.2 } & {27.6 } \\
			{IDC} & {53.7 } & {30.2 } & {33.1 } & {52.2 } & {34.6 } & {47.0 } \\
			{FreD} & {54.2 } & {31.2 } & {32.5 } & {49.1 } & {34.0 } & {43.1 } \\
			{LatentDD} & {56.1 } & {28.0 } & {30.7 } & {—} & {36.3 } & {47.1 } \\
			{DDiF} & {67.8 } & {39.6 } & {43.2 } & {63.1 } & {44.8 } & {67.0 } \\
			{GSDD} & {\textbf{70.0 }} & {\textbf{42.6 }} & {\textbf{51.2 }} & {\textbf{67.4 }} & {\textbf{46.4 }} & {\textbf{70.4 }} \\
			\bottomrule
			\end{tabular}%
			}
		}
	\end{minipage}
	\hfill
	\begin{minipage}[t]{0.355\textwidth}
		\centering  \caption{Cross-Architecture Results across AlexNet, VGG11, ResNet18 and ViT.}
		\label{tab:cross-arch}
		\resizebox{\linewidth}{!}{
			\setlength{\tabcolsep}{1.5pt}
			\begin{tabular}{ccccccc}    \toprule    Subset & Nette & Woof  & Fruit & Yellow & Meow  & Squawk \\    \midrule    TM    & 22.0    & 14.8  & 17.1  & 22.3  & 16.2  & 25.5 \\    IDC   & 27.9  & 19.5  & 23.9  & 28.0    & 19.8  & 29.9 \\    FreD  & 36.2  & 23.7  & 23.6  & 31.2  & 19.1  & 37.4 \\    GLaD  & 30.4  & 17.1  & 21.1  & —     & 19.6  & 28.2 \\    H-GLaD & 30.8  & 17.4  & 21.5  & —     & 20.1  & 28.8 \\    LD3M  & 32.0    & 19.9  & 21.4  & —     & 22.1  & 30.4 \\    DDiF  & \textbf{59.3} & 34.1  & 39.3  & 51.1  & 33.8  & 54.0 \\    GSDD  & 58.1  & \textbf{34.6} & \textbf{39.9} & \textbf{53.6} & \textbf{34.0} & \textbf{58.0} \\    \bottomrule
			\end{tabular}
		}
	\end{minipage}
	\hfill
	 \begin{minipage}[t]{0.20\textwidth}
	\centering \caption{Ablation Study on ImageNette for $128\times 128$.}
	\label{tab:ablation}%
	\resizebox{\linewidth}{!}{
		\setlength{\tabcolsep}{2pt}
		\renewcommand{\arraystretch}{1.13}
		{
		\begin{tabular}{cc}
		\toprule
		Module & Acc \\
		\midrule
		naive & 72.8±1.5 \\
		w/ bf16  & 73.2±1.2 \\
		w/ boundary & 73.0±0.9 \\
		w/ opacity & 73.5±1.6 \\
		w/ antialias & 73.3±1.0 \\
		w/ all   & 74.4±1.2 \\
		\bottomrule
		\end{tabular}%
		}
	}
	 \end{minipage}
\end{table*}

\subsection{Experimental Results}

\paragraph{Datasets and Baselines} We evaluate our method on standard datasets: CIFAR-10, CIFAR-100 (at $32 \times 32$ resolution), six ImageNet subsets (at $128 \times 128$ resolution and $256 \times 256$ resolution), and ImageNet-1K (at $224\times 224$ resolution). The general pipeline of dataset distillation involves two stages: (1) generating a synthetic dataset, and (2) training a model from scratch on the synthetic data. The test accuracy of the trained model is used to assess the quality of the distilled dataset. We include a wide range of baselines, including TM~\cite{Cazenavette_2022_CVPR}, FRePo~\cite{zhouDatasetDistillationUsing2022}, IDC~\cite{kimDatasetCondensationEfficient2022}, FreD~\cite{shinFrequencyDomainBasedDataset2023}, HaBa~\cite{liuDatasetDistillationFactorization2022}, RTP~\cite{dengRememberDistillingDatasets2022}, HMN~\cite{liuDatasetDistillationFactorization2022}, SPEED~\cite{weiSparseParameterizationEpitomic2023}, NSD~\cite{NSD}, GLaD~\cite{cazenavetteGeneralizingDatasetDistillation2023}, H-GLaD~\cite{Zhong_2025_CVPR}, LD3M~\cite{moser2024latentdatasetdistillationdiffusion} and DDiF~\cite{shin2025distilling}.
\vspace{-0.1cm}

\paragraph{Experimental Details} We use TM~\cite{Cazenavette_2022_CVPR}, DM~\cite{zhaoDatasetCondensationDistribution2023}, and DC~\cite{zhaoDatasetCondensationGradient2021} as the underlying distillation algorithms in our experiments. All parameters of Gaussians are trained using the Adam optimizer, and the learning rate for Gaussian parameters is uniformly set to 0.001. For CIFAR datasets, we apply ZCA whitening as a standard preprocessing step. We fix the boundary loss weight to $\lambda_{\text{boundary}} = 0.1$. Experiments are conducted on NVIDIA V100, A100 and RTX 4090 GPUs. As a default preprocessing in dataset parameterization~\cite{kimDatasetCondensationEfficient2022,shinFrequencyDomainBasedDataset2023,shin2025distilling}, we use real image sampled randomly from original dataset to initialize the representation except for ImageNet-1K for a fair comparison with SRe2L~\cite{yinSqueezeRecoverRelabel2023}. We provide complete details and hyperparameters in the Appendix~\ref{details}.

\paragraph{Storage Budget and GPC} We evaluate performance under varying storage budgets, specified in IPC (images per class). Since our method utilizes multiple low-storage Gaussian images, we additionally report GPC (Gaussian Images per Class) in Appendix~\ref{details}. The number of Gaussian points per image is computed as $\text{pts} = \frac{\text{res} \times \text{res} \times 3 \times \text{IPC} \times 2}{\text{GPC} \times 9}$, where the factor of 2 accounts for the use of bf16, and the denominator 9 reflects the number of parameters per Gaussian.

\paragraph{Performance Comparison} Through the integration of GSDD, our method consistently outperforms the pixel-based TM baseline and other dataset parameterization methods on six ImageNet subsets with IPC=1, as shown in Table~\ref{tab:subset-mtt-ipc1}.
In addition, we also report experiments conducted at a lower resolution on CIFAR-10/100 with IPC=1,10,50, with the results presented in Appendix Table~\ref{tab:cifar-mtt-ipc1}. These results highlight the effectiveness of GSDD in enhancing the quality of distilled datasets.
Given the efficiency of GSDD, we additionally report performance on higher-resolution ImageNet subset (256×256) in Table~\ref{tab:subset-dm256-ipc1} and large scale dataset ImageNet-1K in Table~\ref{tab:imagenet_1k}. To the best of our knowledge, this is the first time that dataset parameterization techniques used in ImageNet-1K, and we additionally reproduce the IDC (Resize) performance on ImageNet-1K. Our method remains superior at higher resolutions and larger dataset. We use warmup initialization strategy instead of using real image for a fair comparison, which is detailed in Appendix~\ref{imagenet1k_detail}.

\begin{table}[t]
	\centering
	\begin{minipage}[b]{0.61\linewidth}
	\centering \caption{\small Comparison of Parameterization Methods on Image Fitting Efficiency and Representational Quality.}
	\resizebox{\linewidth}{!}{
	\setlength{\tabcolsep}{2pt}
	\renewcommand{\arraystretch}{0.8}
    \begin{tabular}{cccc}
    \toprule
    \multirow{2}[2]{*}{Method } & PSNR  &  Peak Mem &  Total Time \\
          & (dB)  & (MB)  & (s) \\
    \midrule
    DDiF  & 23.21  & 1078.8 & 2206.1  \\
    FreD  & 15.69  & 280.4 & 34.0  \\
    GSDD  & 23.35  & 150.5 & 49.7  \\
    SPEED & 18.05  & 298.5 & 187.1  \\
    \bottomrule
    \end{tabular}%
	}
	\label{tab:benchmark_psnr}%
	\end{minipage}
	\begin{minipage}[b]{0.378\linewidth}
	\caption{\small Performance on ImageNet-1K based on SRe2L~\cite{yinSqueezeRecoverRelabel2023}.}
	\resizebox{\linewidth}{!}{
	\setlength{\tabcolsep}{2pt}
	\renewcommand{\arraystretch}{0.8}
    \begin{tabular}{ccc}
    \toprule
    \multirow{3}[4]{*}{Method} & \multicolumn{2}{c}{ImageNet-1K} \\
          & \multicolumn{2}{c}{($224\times 224$)} \\
\cmidrule{2-3}          & IPC=1 & IPC=10 \\
    \midrule
    SRe2L & 0.9   & 21.3 \\
    +IDC  & 13.4  & 41.5 \\
    +GSDD & 29.0    & 46.2 \\
    \bottomrule
    \end{tabular}%
	}
	\label{tab:imagenet_1k}%
	\end{minipage}
	\vspace{-0.7cm}
\end{table}

\FloatBarrier
\begin{figure*}[t]
	\centering
	\begin{subfigure}[b]{0.24\textwidth}
	\includegraphics[width=\textwidth]{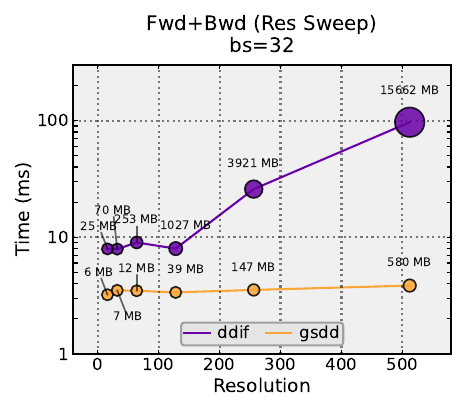}
	\caption{\small Resolution Scaling.}
	\end{subfigure}
	\begin{subfigure}[b]{0.24\textwidth}
	\includegraphics[width=\textwidth]{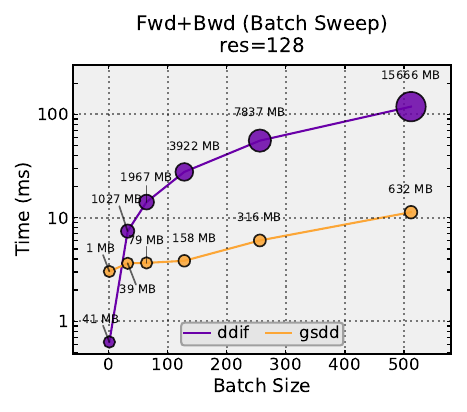}
	\caption{\small Batch-size Scaling.}
	\end{subfigure}
	\begin{subfigure}[b]{0.24\textwidth}
	\includegraphics[width=\textwidth]{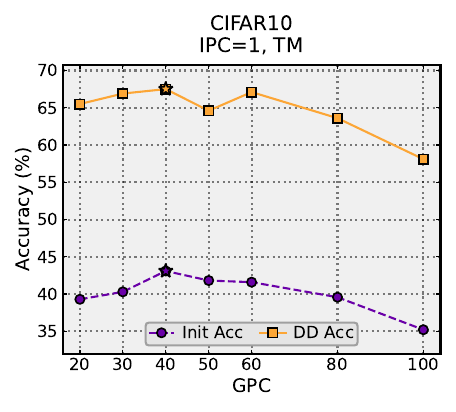}
	\caption{\small GPC scaling (CIFAR-10).}
	\end{subfigure}
	\begin{subfigure}[b]{0.24\textwidth}
	\includegraphics[width=\textwidth]{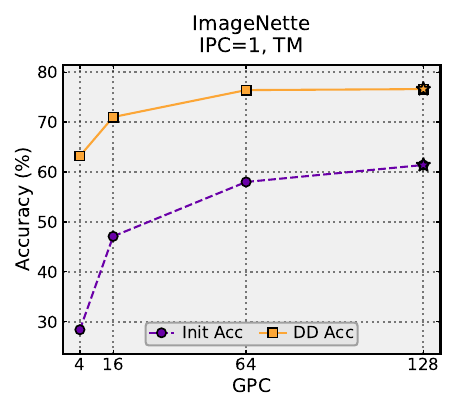}
	\caption{\small GPC scaling (ImageNette)}
	\end{subfigure}
	\caption{
	\textbf{Efficiency, scalability, and performance scaling of GSDD.}
		\textbf{(a)} Execution time and peak GPU memory consumption as a function of batch size for GSDD and DDiF at a fixed resolution of 128. 
		\textbf{(b)} Computational overhead scaling with image resolution at a fixed batch size of 32. 
		\textbf{(c, d)} Accuracy comparison between only initialization (Init Acc) and distilled results (DD Acc) across varying Gaussian images per class (GPC) under a fixed storage budget.
	}
	\label{fig:efficiency_and_ablation}
	\vspace{-0.2cm}
\end{figure*}

\paragraph{Universality to Matching Objectives} GSDD is designed to be compatible with a wide range of dataset distillation algorithms and can be directly integrated with them. We evaluate GSDD under two widely adopted loss paradigms: gradient matching (DC) and distribution matching (DM). Results shown in Table~\ref{tab:subset-dm-dc-ipc1} confirm that GSDD robustly enhances performance across different distillation objectives.

\paragraph{Video Performance}
Benefiting from structural simplicity and high computational efficiency, we further evaluate GSDD on the task of video dataset distillation. We adopt a setup similar to VD~\cite{wangDancingStillImages2024}, with details provided in Appendix~\ref{details}, and use DM as the default distillation algorithm. The results in Table~\ref{tab:video_result} reveal the substantial redundancy inherent in video data and highlight the potential of data parameterization for video dataset distillation.

\paragraph{Cross-Architecture Performance}
The distilled data should retain high performance even when the evaluation model differs from the one used during distillation. To assess this, we synthesize data using a ConvNet and evaluate it on a variety of downstream architectures, including ResNet, VGG, AlexNet and ViT. The results, summarized in Table~\ref{tab:cross-arch} and detailed in Table~\ref{exp:detailed_cross_arch}, demonstrate great cross-architecture generalization of GSDD.

\subsection{Benchmark Rendering Efficiency}



To evaluate the efficiency and representational capacity of GSDD, we conduct an image-fitting experiment comparing several recent parameterization methods with the same storage budget, including FreD~\cite{shinFrequencyDomainBasedDataset2023}, SPEED~\cite{weiSparseParameterizationEpitomic2023} and DDiF~\cite{shin2025distilling}. Specifically, we use the ImageNette dataset and fit the first 128 images from each class (1,280 images in total) at a resolution of 128. The optimization is performed for 2,000 steps, and for each method we perform a grid search over learning rates and optimizers to ensure a fair comparison.
Experiment details is shown in Appendix~\ref{performance_benchmarking}.
As shown in Table \ref{tab:benchmark_psnr}, GSDD shows great performance across representational quality (PSNR), GPU memory usage, and runtime efficiency. High fidelity and low computational overhead of GSDD enables fast synthesis of more diverse (Higher GPC) distilled datasets, which in turn leads to improved performance.

We further conduct a comprehensive quantitative comparison with the state-of-the-art DDiF method in terms of both decoding and training performance. As shown in Figure~\ref{fig:efficiency_and_ablation}(a) and (b), we measure the inference time and memory consumption of both methods across varying per-image storage budgets, image resolutions, and batch sizes.
The results demonstrate that GSDD consistently outperforms DDiF in both inference/training speed and memory efficiency. This advantage becomes especially pronounced when handling high-resolution inputs or large batch sizes, where GSDD achieves several-fold, or even an order-of-magnitude, reductions in both computation time and memory usage.
These findings indicate that our method not only maintains strong image representation capabilities for distillation but also offers superior rendering speed and lower memory footprint, thereby substantially improving the scalability of dataset distillation methods in practical deployment scenarios.

\subsection{Ablation Study and Additional Experiments}
\label{subsection:ablation_study}

Table~\ref{tab:ablation} summarizes the ablation results for the components detailed in Section~\ref{subsec:design}, confirming that each element is essential for achieving optimal performance.
We further examine the effect of GPC through an ablation study, as shown in Figure~\ref{fig:efficiency_and_ablation}, and more experimental results is shown in Figure~\ref{fig:gpc_ablation_appendix} in the Appendix.
When distilling under small storage budgets as shown in Figure~\ref{fig:efficiency_and_ablation}(c), performance increases with GPC at first and then declines. This is because small GPC values lead to insufficient diversity, whereas excessively large GPC values allocate too few Gaussians per image, weakening each image's representational capacity. In more memory-consuming settings as shown in Figure~\ref{fig:efficiency_and_ablation}(d), the optimal turning point of GPC becomes difficult to reach, and performance tends to improve monotonically. Figure~\ref{fig:efficiency_and_ablation} also shows that the performance of the initial dataset serves as a useful reference for estimating a reasonable GPC range.

Additional experiments are provided in the Appendix, including detailed cross-architecture evaluations in Table~\ref{exp:detailed_cross_arch}, efficiency benchmark of GSDD in Table~\ref{fig:sweep_all}, ablations on initialization strategies in Table~\ref{tab:init_ablation}, comprehensive GPC ablations in Figure~\ref{fig:gpc_ablation_appendix}, stability analyses of the CUDA implementation across different GPU architectures in Table~\ref{tab:benchmark_arch}, cross-resolution evaluations in Table~\ref{tab:cross_resolution}, as well as further visualizations in Appendix~\ref{visualization_antialias_escape} and Appendix~\ref{visualization_gsdd}.

\section{Conclusion}

This work introduces sparse Gaussian representations into dataset distillation parameterization and proposes a unified framework for encoding distilled images using 2D Gaussian primitives.
Through extensive quantitative and qualitative evaluations, we demonstrate that GSDD combine strong representational capacity with high efficiency, enabling more diverse distilled datasets and improved performance, and extension to high computation scenarios.
Future work may explore dynamic density control of Gaussian primitives to improve representational efficacy.


\bibliography{paper}
\bibliographystyle{icml2026}

\newpage
\appendix
\onecolumn
\appendix
\section{Dataset Distillation Objectives}
\label{related_works}

Dataset distillation treats synthetic data as learnable parameters. The primary optimization objective can be formulated as:
\begin{equation}
\begin{aligned}
\mathcal{S}^* = \arg\min_{\mathcal{S}} & \left| \mathcal{L}(\theta_{\mathcal{S}}, \mathcal{D}) - \mathcal{L}(\theta_{\mathcal{T}}, \mathcal{D}) \right| \\
\text{s.t. } \theta_{\mathcal{S}} = \arg\min_{\theta} & \mathcal{L}(\theta, \mathcal{S}), \quad \theta_{\mathcal{T}} = \arg\min_{\theta} \mathcal{L}(\theta, \mathcal{D}),
\end{aligned}
\end{equation}
and an empirical implementation could be formulates in the meta-learning paradigm~\cite{wangDatasetDistillation2018}:
\begin{equation}
\begin{aligned}
\mathcal{S}^* = \arg\min_{\mathcal{S}} \mathcal{L}(\theta_{\mathcal{S}}, \mathcal{T}),\quad\text{s.t. } \theta_{\mathcal{S}} = \arg\min_{\theta} \mathcal{L}(\theta, \mathcal{S}),
\end{aligned}
\end{equation}
where $\mathcal{D,S,T}$ denote the underlying data distribution, distilled dataset, and original dataset respectively. Such meta-learning paradigm may incur substantial computational costs and face challenges such as vanishing gradients or weak supervision signals. Consequently, many alternative objectives have been proposed to align the training-aware information between original dataset and distilled dataset.

\paragraph{Trajectory Matching (TM)}
Trajectory Matching~\cite{Cazenavette_2022_CVPR,duMinimizingAccumulatedTrajectory2023,liuDatasetDistillationAutomatic2025,guoLosslessDatasetDistillation2024,cuiScalingDatasetDistillation2023} aims to align the parameter evolution of a student model trained on the synthetic set with the trajectory of an expert model trained on real data:
\begin{equation}
\mathcal{S}^* = \arg\min_{\mathcal{S}} \mathbb{E}_{\theta_{\mathcal{T}}^i} \left\| \hat{\theta}_{\mathcal{S}}(\theta_{\mathcal{T}}^i, N) - \theta_{\mathcal{T}}^{i+M} \right\|_2^2,
\end{equation}
where $\theta_{\mathcal{T}}^i$ denotes the parameter state of an expert model pre-trained on $\mathcal{D}$ at the $i$-th epoch, and $\hat{\theta}_{\mathcal{S}}(\theta_{\mathcal{T}}^i, N)$ represents the state of a student model initialized at $\theta_{\mathcal{T}}^i$ after $N$ update steps on the synthetic set $\mathcal{S}$. The number of steps $N$ should be carefully chosen to ensure the student trajectory aligns with the target expert state $\theta_{\mathcal{T}}^{i+M}$~\cite{liuDatasetDistillationAutomatic2025}.

\paragraph{Distribution Matching (DM)}
Distribution Matching~\cite{zhaoDatasetCondensationDistribution2023,zhaoImprovedDistributionMatching2023,liuDREAMEfficientDataset2023a,Wei_2024_CVPR,Wang_2025_CVPR} focuses on aligning the statistical properties of the synthetic and original data within a shared feature space:
\begin{equation}
\mathcal{S}^* = \arg\min_{\mathcal{S}} \sum_{k=1}^K \left\| \mathbb{E}_{s \sim \mathcal{S}_k} [\psi_\theta(s)] - \mathbb{E}_{t \sim \mathcal{T}_k} [\psi_\theta(t)] \right\|_2^2,
\end{equation}
where $\psi_\theta(\cdot)$ denotes a feature embedding extracted by a network with parameters $\theta$, and $k$ denotes the class index. This objective minimizes the distance between the class-wise mean embeddings of synthetic and real samples, offering high computational efficiency by avoiding nested optimization loops.

\paragraph{Gradient Matching (DC/GM)}
Gradient Matching~\cite{zhaoDatasetCondensationGradient2021,leeDatasetCondensationContrastive2022,duuentialSubsetMatching2023} ensures that the model gradients derived from the synthetic set align with those generated by the original dataset:
\begin{equation}
\mathcal{S}^* = \arg\min_{\mathcal{S}} \mathbb{E}_{\theta \sim P_\theta} \left[ \text{Dist}\left(\nabla_\theta \mathcal{L}(\theta, \mathcal{S}), \nabla_\theta \mathcal{L}(\theta, \mathcal{T})\right) \right],
\end{equation}
where $\text{Dist}(\cdot, \cdot)$ is a distance metric such as cosine similarity or the $L_2$ norm. By matching gradients at each training iteration, the synthetic data is optimized to induce learning dynamics similar to those of the full dataset.

\paragraph{Decoupled Matching}
Decoupled distillation, introduced by SRe2L~\cite{yinSqueezeRecoverRelabel2023} and subsequently extended by various methods such as EDC~\cite{shao2024elucidating} and GIFT~\cite{shang2025gift}, decomposes the distillation process into two independent stages: model pre-training and data synthesis. 
These approaches utilize pre-trained teacher models to recover synthetic samples by aligning internal feature statistics, thereby bypassing the computationally expensive nested optimization loops. 
Formally, the synthesis objective is formulated as:
\begin{equation}
\mathcal{S}^* = \arg\min_{\mathcal{S}} \sum_{l=1}^L \left( \left\| \mu_l(\mathcal{S}) - \hat{\mu}_l \right\|_2^2 + \left\| \sigma_l^2(\mathcal{S}) - \hat{\sigma}_l^2 \right\|_2^2 \right),
\end{equation}
where $\mu_l(\mathcal{S})$ and $\sigma_l^2(\mathcal{S})$ represent the mean and variance of the feature maps at the $l$-th layer of the teacher model $\theta_{\mathcal{T}}$ when processing synthetic data $\mathcal{S}$, while $\hat{\mu}_l$ and $\hat{\sigma}_l^2$ denote the batch normalization (BN) statistics pre-stored in the teacher model, which serve as anchors for knowledge recovery. 
Here, $L$ signifies the total number of layers selected for alignment. 
By further incorporating data augmentation techniques such as CutMix~\cite{Yun_2019_ICCV} and employing soft labels, these methods have achieved state-of-the-art performance in large-scale dataset distillation tasks.

\section{Experimental Details}
\label{details}

\paragraph{Datasets}

We evaluate the proposed method on a variety of benchmarks, encompassing both image and video datasets.
For \textbf{image dataset distillation}, we utilize CIFAR-10 and CIFAR-100 at $32 \times 32$ resolution. Both datasets consist of 60,000 images, with 50,000 for training and 10,000 for testing, distributed across 10 and 100 classes respectively. We further incorporate six ImageNet subsets (\textit{imagenette}, \textit{imagewoof}, \textit{imagefruit}, \textit{imagemeow}, \textit{imagesquawk}, and \textit{imageyellow}), each containing 10 classes and over 10,000 images, evaluated at resolutions of $128 \times 128$ and $256 \times 256$. Finally, experiments are scaled to the full ImageNet-1K benchmark at $224 \times 224$ resolution.
For \textbf{video dataset distillation}, we follow the experimental protocols established in VD~\cite{wangDancingStillImages2024}. We employ MiniUCF and HMDB51, which contain 50 and 51 classes, respectively. In line with previous work, video frames are cropped and resized to $112 \times 112$. We sample 16-frame clips using a temporal stride of 4.

\paragraph{Network Architectures}
We adopt ConvNet architectures as the default backbone networks. A Depth-$n$ ConvNet consists of $n$ blocks followed by a fully-connected layer, where each block is composed of a $3 \times 3$ convolutional layer with 128 filters, instance normalization, a ReLU activation, and $2 \times 2$ average pooling with stride 2. 
Specifically, we use ConvNetD3 for CIFAR-10 and CIFAR-100, ConvNetD5 for the ImageNet subsets, ResNet18 for ImageNet-1K, ConvNet3D for MinUCF and HMDB51.
Specific implementation can be checked in the code.

\paragraph{Distilled Dataset Initialization} For the initialization of distilled data, all experiments adopt the MSE loss and use the Adam optimizer with a learning rate of $1\times10^{-2}$ to reconstruct the randomly sampled images from the real training set.

\paragraph{Distillation Details}
During the distillation process, we optimize the Gaussian point attributes using the Adam optimizer. All attributes share a uniform learning rate of $1\times10^{-3}$, except for the $xy$ coordinates in video distillation, which remain fixed with a learning rate of zero. Unless otherwise specified, the optimization procedure spans 15,000 iterations for Trajectory Matching (TM), 20,000 for Decoupled Matching (DM), and 2,000 for Distribution Matching (DC), respectively. Hyperparameters for different setup has been detailed in Table~\ref{tab:cifar10tm},\ref{tab:cifar100tm},\ref{tab:subsetstm},\ref{tab:subsetsdm},\ref{tab:subsetsdm256} and \ref{tab:subsetsdc}.

\paragraph{Scaling to ImageNet-1K}
\label{imagenet1k_detail}

\begin{wraptable}{r}{0.35\linewidth}
	\centering
	\caption{\small Hyperparameters on ImageNet-1K.}
	\scalebox{0.9}{
	\setlength{\tabcolsep}{6pt}
	\begin{tabular}{ccc}
	\toprule
		& IPC=1 & IPC=10 \\
	\midrule
	SRe2L+IDC   & k=20   & k=5 \\
	SRe2L+GSDD  & GPC=50 & GPC=200 \\
	\bottomrule
	\end{tabular}
	}
	\label{tab:imagenet_1k_hyper}%
\end{wraptable}

Benefiting from the efficiency of GSDD and efficiency of recent dataset-distillation frameworks, such as SRe2L~\cite{yinSqueezeRecoverRelabel2023} and RDED~\cite{sunDiversityRealismDistilled2024}, we extend our evaluation to the large-scale ImageNet-1K benchmark~\cite{5206848}. 
We adopt SRe2L~\cite{yinSqueezeRecoverRelabel2023} as the primary distillation framework due to its established effectiveness in large-scale scenarios. 
Furthermore, we implement the IDC baseline~\cite{kimDatasetCondensationEfficient2022} within the SRe2L pipeline, which parameterizes condensed data through a resizing technique. In IDC, each image is spatially downscaled by a factor of $\sqrt{k}$, resulting in $k$-times more synthetic samples.
To ensure a competitive baseline, we perform a grid search for the optimal resize factor $k$ for IDC, considering $k \in \{4, 20, 50\}$ for IPC=1 and $k \in \{4, 10, 20\}$ for IPC=10.
To maintain a fair comparison with SRe2L's random initialization protocol, GSDD employs a warmup initialization strategy rather than initializing from real images. 
We found optimizing Gaussian points from a purely random start can lead to instability. 
A detailed analysis of various initialization strategies is provided in Appendix~\ref{ablation_initialization}. 
The final configurations for the resize ratio $k$ in IDC and the Gaussian point counts in GSDD are summarized in Table~\ref{tab:imagenet_1k_hyper}.

\paragraph{Video Dataset Distillation}
We represent each video as a sequence of frames, where each frame is parameterized by a set of Gaussian primitives. We adopt DM as the distillation algorithm due to its computational efficiency.
For the baseline that directly trains ConvNet3D on the video dataset, each video consists of 16 frames sampled with a frame interval of four.
During both training and testing, the starting frame is dynamically selected. This differs slightly from the implementation in VD~\cite{wangDancingStillImages2024}, where the starting frame is randomly selected in the first iteration and then fixed throughout training, which may be an implementation artifact. During evaluation, we sample 10 different starting positions and report the averaged results to ensure robustness.
For both full-dataset training and the final evaluation on distilled datasets produced by GSDD, the number of training epochs is set to 1,000. The learning rate is fixed to 1e-2, and SGD is used as the optimizer.

\begin{table}[h]
	\centering
	\caption{Hyperparameters for dataset distillation on CIFAR-10 (TM) $32\times32$.}
	\scalebox{0.8}{
	\label{tab:cifar10tm}
	\begin{tabular}{lccccccccc}
		\toprule
		Setting & GPC & num\_points & syn\_steps & max\_start\_epoch & expert\_epochs & lr\_lr & lr\_init & batch\_syn & zca \\
		\midrule
		IPC=1  & 30  & 22  & 60 & 30 & 2 & 1e-5 & 1e-2 & 300 & True \\
		IPC=10 & 160 & 42  & 60 & 30 & 2 & 1e-5 & 1e-2 & 380 & True \\
		IPC=50 & 250 & 136 & 60 & 30 & 2 & 1e-5 & 1e-2 & 360 & True \\
		\bottomrule
	\end{tabular}
}

	\centering
	\caption{Hyperparameters for dataset distillation on CIFAR-100 (TM) $32\times32$.}
		\scalebox{0.8}{
	\label{tab:cifar100tm}
	\begin{tabular}{lccccccccc}
		\toprule
		Setting & GPC & num\_points & syn\_steps & max\_start\_epoch & expert\_epochs & lr\_lr & lr\_init & batch\_syn & zca \\
		\midrule
		IPC=1        & 30  & 22 & 60 & 30 & 2 & 1e-5 & 1e-2 & 512 & True \\
		IPC=10       & 80  & 85 & 60 & 30 & 2 & 1e-5 & 1e-2 & 512 & True \\
		IPC=50 & 400 & 85 & 60 & 30 & 2 & 1e-5 & 1e-2 & 640 & True \\
		\bottomrule
	\end{tabular}
}

	\centering
	\caption{Hyperparameters for dataset distillation on ImageNet-subset (TM) $128\times128$.}
		\scalebox{0.8}{
	\label{tab:subsetstm}
	\begin{tabular}{lccccccccc}
		\toprule
		Setting & GPC & num\_points & syn\_steps & max\_start\_epoch & expert\_epochs & lr\_lr & lr\_init & batch\_syn & zca \\
		\midrule
		IPC=1 & 64  & 170 & 20 & 40 & 2 & 1e-6 & 1e-2 & 150 & False \\
		IPC=10      & 640 & 170 & 20 & 40 & 2 & 1e-5 & 1e-2 & 160 & False \\
		\bottomrule
	\end{tabular}
}
\end{table}

\begin{table}[htbp]
	\centering
	\caption{Hyperparameters for dataset distillation on ImageNet-subset $128\times128$ (DM).}
		\scalebox{0.8}{
	\label{tab:subsetsdm}
	\begin{tabular}{lcccccc}
		\toprule
		Setting & GPC & num\_points & batch\_real & batch\_syn & zca & Iteration \\
		\midrule
		IPC=1 & 200 & 54 & 1024 & 2000 & False & 20000 \\
		\bottomrule
	\end{tabular}
}

	\centering
	\caption{Hyperparameters for dataset distillation on ImageNet-subset $256\times256$ (DM).}
		\scalebox{0.8}{
	\label{tab:subsetsdm256}
	\begin{tabular}{lcccccc}
		\toprule
		Setting & GPC & num\_points & batch\_real & batch\_syn & zca & Iteration \\
		\midrule
		IPC=1 & 120 & 364 & 512 & 512 & False & 20000 \\
		\bottomrule
	\end{tabular}
}

	\centering
	\caption{Hyperparameters for dataset distillation on ImageNet-subset $128\times128$ (DC).}
		\scalebox{0.8}{
	\label{tab:subsetsdc}
	\begin{tabular}{lcccccc}
		\toprule
		Setting & GPC & num\_points & batch\_real & batch\_syn & zca & Iteration \\
		\midrule
		IPC=1 & 200 & 54 & 720 & 2000 & False & 2000 \\
		\bottomrule
	\end{tabular}
}
\end{table}



\paragraph{Performance Benchmarking}
\label{performance_benchmarking}

In Table~\ref{tab:benchmark_psnr}, we evaluate the image-fitting performance of FreD~\cite{shinFrequencyDomainBasedDataset2023}, SPEED~\cite{weiSparseParameterizationEpitomic2023}, DDiF~\cite{shin2025distilling}, and GSDD. We select 128 images from each class of the ImageNette dataset, resulting in a total of 1,280 images, all at a resolution of 128. Each method is constrained to the same storage budget, meaning that, on average, each image is represented using 963 floating-point parameters.
For SPEED, we follow the parameter-counting formula provided in its original paper,
$$\text{\#Params}=DK+1.5NHk+R(3D^22+7D)+L(D+1),$$
and, given the storage constraint, we solve for all feasible hyperparameter combinations. The final configuration used in our experiments is (D=32, N=32, k=135, R=2, L=64). We attempted to evaluate HaBa~\cite{liuDatasetDistillationFactorization2022} as well, but its model structure cannot be solved inversely under the resolution of 128 and the storage budget of 963 parameters.
FreD, DDiF, and GSDD exhibit clear advantages in terms of estimating and meeting the storage budget. For each method, we search over Adam and SGD optimizers and learning rates spaced exponentially by a factor of 10, and we report the highest PSNR achieved. All models are trained for 2,000 iterations using MSE loss. Each experiment is repeated three times with random initialization, and the average result is reported.

\section{Additional Experimental Results}

\subsection{Performance Comparison on Low-Dimensional Datasets}

As shown in Table~\ref{tab:cifar-mtt-ipc1}, our method GSDD consistently achieves the improved performance across both CIFAR-10 and CIFAR-100 under all IPC settings. Notably, GSDD surpasses all baselines by a clear margin in the low-data regime (IPC=1), achieving 67.6\% on CIFAR-10 and 43.0\% on CIFAR-100. In high-data settings (IPC=50), GSDD maintains its superiority, reaching 77.7\% and 53.1\% respectively. These results demonstrate the strong generalization ability and scalability of our primitive-based distillation framework.

\begin{table}[htbp]
	\centering  \caption{Classification accuracy on CIFAR-10 and CIFAR-100 under different IPC settings.}
	\label{tab:cifar-mtt-ipc1}
	\resizebox{\textwidth}{!}{
		\begin{tabular}{llcccccc}
			\toprule
			& \multirow{2}{*}{Method} & \multicolumn{3}{c}{CIFAR10} & \multicolumn{3}{c}{CIFAR100} \\
			\cmidrule(lr){3-5} \cmidrule(lr){6-8}
			&  & IPC=1 & IPC=10 & IPC=50 & IPC=1 & IPC=10 & IPC=50 \\
			\midrule
			Input sized & TM    & 46.3±0.8 & 65.3±0.7 & 71.6±0.2 & 24.3±0.3 & 40.1±0.4 & 47.7±0.2 \\
			& FRePo & 46.8±0.7 & 65.5±0.4 & 71.7±0.2 & 28.7±0.1 & 42.5±0.2 & 44.3±0.2 \\
			\midrule
			Static      & IDC   & 50.0±0.4 & 67.5±0.5 & 74.5±0.2 & —        & —        & —        \\
			& FreD  & 60.6±0.8 & 70.3±0.3 & 75.8±0.1 & 34.6±0.4 & 42.7±0.2 & 47.8±0.1 \\
			\midrule
			Parameterized & HaBa  & 48.3±0.8 & 69.9±0.4 & 74.0±0.2 & —        & —        & 47.0±0.2 \\
			& RTP   & 66.4±0.4 & 71.2±0.4 & 73.6±0.5 & 34.0±0.4 & 42.9±0.7 & —        \\
			& HMN   & 65.7±0.3 & 73.7±0.1 & 76.9±0.2 & 36.3±0.2 & 45.4±0.2 & 48.5±0.2 \\
			& SPEED & 63.2±0.1 & 73.5±0.2 & 77.7±0.4 & 40.4±0.4 & 45.9±0.3 & 49.1±0.2 \\
			& NSD   & \textbf{68.5±0.8} & 73.4±0.2 & 75.2±0.6 & 36.5±0.3 & 46.1±0.2 & — \\
			\midrule
			Function    & DDiF  & 66.5±0.4 & 74.0±0.4 & 77.5±0.3 & 42.1±0.2 & 46.0±0.2 & 49.9±0.2 \\
			\midrule
			Primitive   & GSDD  & 67.6±0.4 & \textbf{75.5±0.3} & \textbf{77.7±0.5} & \textbf{43.0±0.1} & \textbf{47.4±0.3} & \textbf{53.1±0.2} \\
			\bottomrule
		\end{tabular}
	}
\end{table}
\subsection{Detailed Cross-architecture Results}
We evaluate the generalization ability of distilled data across different architectures, including AlexNet, VGG11, ResNet18, and ViT. The network implementations follow those provided in~\cite{cazenavetteGeneralizingDatasetDistillation2023}. For training, the learning rate is set to $1\times10^{-3}$ for AlexNet, $5\times10^{-5}$ for ViT, and $1\times10^{-2}$ for the remaining architectures. All models are optimized using Adam with a cosine annealing learning rate schedule. We report the results averaged over three independent runs, as shown in Table~\ref{exp:detailed_cross_arch}.

\begin{table}[t]
	\centering
	\caption{Detailed cross-architecture performance comparison. }
	\label{exp:detailed_cross_arch}
	\resizebox{\textwidth}{!}{
    \begin{tabular}{cccccccc}
    \toprule
    Test Net & Method & Nette & Woof  & Fruit & Yellow & Meow  & Squawk \\
    \midrule
    \multirow{5}[2]{*}{AlexNet} & TM    & 13.2 ± 0.6 & 10.0 ± 0.0 & 10.0 ± 0.0 & 11.0 ± 0.2 & 9.8 ± 0.0 & — \\
          & IDC   & 17.4 ± 0.9 & 16.5 ± 0.7 & 17.9 ± 0.7 & 20.6 ± 0.9 & 16.8 ± 0.5 & 20.7 ± 1.0 \\
          & FreD  & 35.7 ± 0.4 & 23.9 ± 0.7 & 15.8 ± 0.7 & 19.8 ± 1.2 & 14.4 ± 0.5 & 36.3 ± 0.3 \\
          & DDiF  & 60.7 ± 2.3 & \textbf{36.4 ± 2.3} & 41.8 ± 0.6 & \textbf{56.2 ± 0.8} & \textbf{40.3 ± 1.9} & \textbf{60.5 ± 0.4} \\
          & GSDD  & \textbf{63.1 ± 1.3} & 33.6 ± 0.9 & \textbf{41.9 ± 2.0} & 53.5 ± 0.9 & 38.1 ± 0.8 & 60.1 ± 0.2 \\
    \midrule
    \multirow{5}[2]{*}{VGG11} & TM    & 17.4 ± 2.1 & 12.6 ± 1.8 & 11.8 ± 1.0 & 16.9 ± 1.1 & 13.8 ± 1.3 & — \\
          & IDC   & 19.6 ± 1.5 & 16.0 ± 2.1 & 13.8 ± 1.3 & 16.8 ± 3.5 & 13.1 ± 2.0 & 19.1 ± 1.2 \\
          & FreD  & 21.8 ± 2.9 & 17.1 1.7 & 12.6 ± 2.6 & 18.2 ± 1.1 & 13.2 ± 1.9 & 18.6 ± 2.3 \\
          & DDiF  & 53.6 ± 1.5 & 29.9 ± 1.9 & 33.8 ± 1.9 & 44.2 ± 1.7 & \textbf{32.0 ± 1.8} & 37.9 ± 1.5 \\
          & GSDD  & \textbf{56.8 ± 1.6} & \textbf{32.7 ± 1.4} & \textbf{34.1 ± 3.0} & \textbf{57.3 ± 4.5} & 31.3 ± 1.8 & \textbf{55.9 ± 3.4} \\
    \midrule
    \multirow{5}[2]{*}{ResNet18} & TM    & 34.9 ± 2.3 & 20.7 ± 1.0 & 23.1 ± 1.5 & 43.4 ± 1.1 & 22.8 ± 2.2 & — \\
          & IDC   & 43.6 ± 1.3 & 23.2 ± 0.8 & 32.9 ± 2.8 & 44.2 ± 3.5 & 28.2 ± 0.5 & 47.8 ± 1.9 \\
          & FreD  & 48.8 ± 1.8 & 28.4 ± 0.6 & 34.0 ± 1.9 & 49.3 ± 1.1 & 29.0 ± 1.8 & 50.2 ± 0.8 \\
          & DDiF  & \textbf{63.8 ± 1.8} & 37.5 ± 1.9 & 42.0 ± 1.9 & \textbf{55.9 ± 1.0} & 35.8 ± 1.8 & \textbf{62.6 ± 1.5} \\
          & GSDD  & 59.1 ± 3.0 & \textbf{39.5 ± 2.0} & \textbf{42.6 ± 0.8} & \textbf{55.9 ± 1.2} & \textbf{40.9 ± 2.9} & \textbf{62.6 ± 0.6} \\
    \midrule
    \multirow{5}[2]{*}{ViT} & TM    & 22.6 ± 1.1 & 15.9 ± 0.4 & 23.3 ± 0.4 & 18.1 ± 1.3 & 18.6 ± 0.9 & — \\
          & IDC   & 31.0 ± 0.6 & 22.4 ± 0.8 & 31.1 ± 0.8 & 30.3 ± 0.6 & 21.4 ± 0.7 & 32.2 ± 1.2 \\
          & FreD  & 38.4 ± 0.7 & 25.4 ± 1.7 & 31.9 ± 1.4 & 37.6 ± 2.0 & 19.7 ± 0.8 & 44.4 ± 1.0 \\
          & DDiF  & \textbf{59.0 ± 0.4} & \textbf{32.8 ± 0.8} & 39.4 ± 0.8 & \textbf{47.9 ± 0.6} & \textbf{27.0 ± 0.6} & \textbf{54.8 ± 1.1} \\
          & GSDD  & 53.4 ± 0.6 & 32.6 ± 1.1 & \textbf{41.0 ± 1.2} & 47.6 ± 0.8 & 25.8 ± 0.4 & 53.5 ± 2.1 \\
    \bottomrule
    \end{tabular}%
	}
\end{table}

\subsection{Additional Ablation Study}

\begin{figure*}[t]
	\centering
	\includegraphics[width=0.99\linewidth]{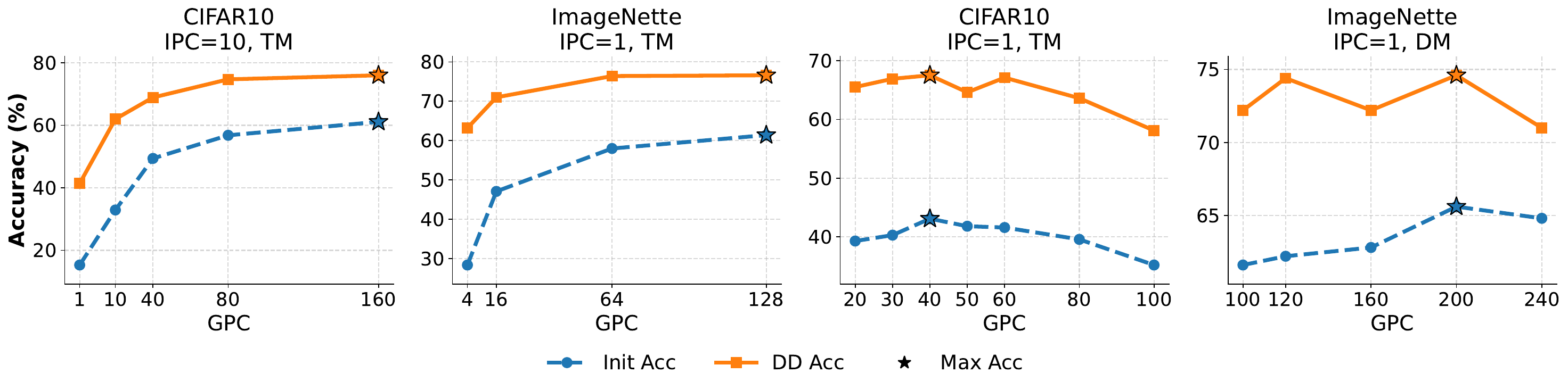}
	\caption{Effect of GPC (Gaussian Images Per Class) on distilled dataset performance across different datasets, storage budgets, and distillation algorithms (TM~\cite{Cazenavette_2022_CVPR} and DM~\cite{zhaoDatasetCondensationDistribution2023}). Initial Accuracy denotes the performance of initialized gaussian images (initialized on sampled real images). Distilled Accuracy denotes the performance after distillation.}
	\label{fig:gpc_ablation_appendix}
\end{figure*}

\begin{wraptable}{r}{0.24\textwidth}
	\centering \caption{\small Initialization Strategy Ablation Results}
	\label{tab:init_ablation}
	\scalebox{0.9}{
	\setlength{\tabcolsep}{1.5pt}
	\renewcommand{\arraystretch}{0.95}
	\begin{tabular}{ccc}
	\toprule
	& Init & Acc. \\
	\midrule
	IDC   & real  & 61.4  \\
	FreD  & real  & 66.8  \\
	NSD   & random & 68.6  \\
	DDiF  & real  & 72.0  \\
	GSDD  & random & 62.2  \\
	GSDD  & warmup & 66.7  \\
	GSDD  & real  & 76.4  \\
	\bottomrule
	\end{tabular}%
	}
\end{wraptable}

\paragraph{Initialization Strategies}
\label{ablation_initialization}
Prior methods often initialize distilled data using real images, which may introduce additional privacy risks. We further conduct an ablation study on different initialization strategies, including random initialization. However, we find that optimizing Gaussian representations from purely random initialization is inherently unstable: the initial Gaussian points are extremely small and sparsely distributed, creating large gaps that prevent effective gradient propagation. To mitigate this issue, we introduce a Solid-Color Warmup strategy. Instead of fitting any real images, all Gaussian representations are first optimized to fit a single solid-color image. This process reduces point sparsity and adjusts the initial point sizes, after which standard distillation begins. As shown in Table~\ref{tab:init_ablation}, even without real-image initialization, our warmup-based strategy achieves strong results and surpasses several methods that depend on real-image initialization.

\paragraph{Ablation on GPC}
We evaluate the performance variation of distilled datasets as a function of the Gaussian images Per Class (GPC) across various experimental settings. 
As illustrated in Figure~\ref{fig:gpc_ablation_appendix}, we conduct tests on CIFAR-10 and ImageNette datasets under both IPC=1 and IPC=10 configurations using Trajectory Matching (TM) and Distribution Matching (DM) algorithms. 
Our observations reveal distinct trends based on the memory footprint of the tasks. 
In memory-intensive scenarios (left two panels), performance consistently improves with increasing GPC before reaching a plateau. 
Conversely, in settings with lower memory overhead (right panels), performance initially increases and then subsequently declines as GPC continues to grow. 
We hypothesize that when memory resources are abundant, an excessively high GPC may lead to the under-representation of individual distilled images, thereby hindering overall performance. 
In contrast, under high-memory demand, GPU constraints typically prevent the GPC from reaching this point of diminishing returns; thus, a larger GPC generally yields better results within the hardware limits. 
Furthermore, we examine the initial performance of the synthetic sets (before distillation). 
We find a positive correlation between the initial and final performance, suggesting that the initial quality of the synthetic dataset can serve as a reliable proxy for estimating the optimal GPC.

\subsection{Performance Benchmark for Parameterization Performance}
\begin{figure*}[htbp]
	\centering
	\includegraphics[width=\linewidth]{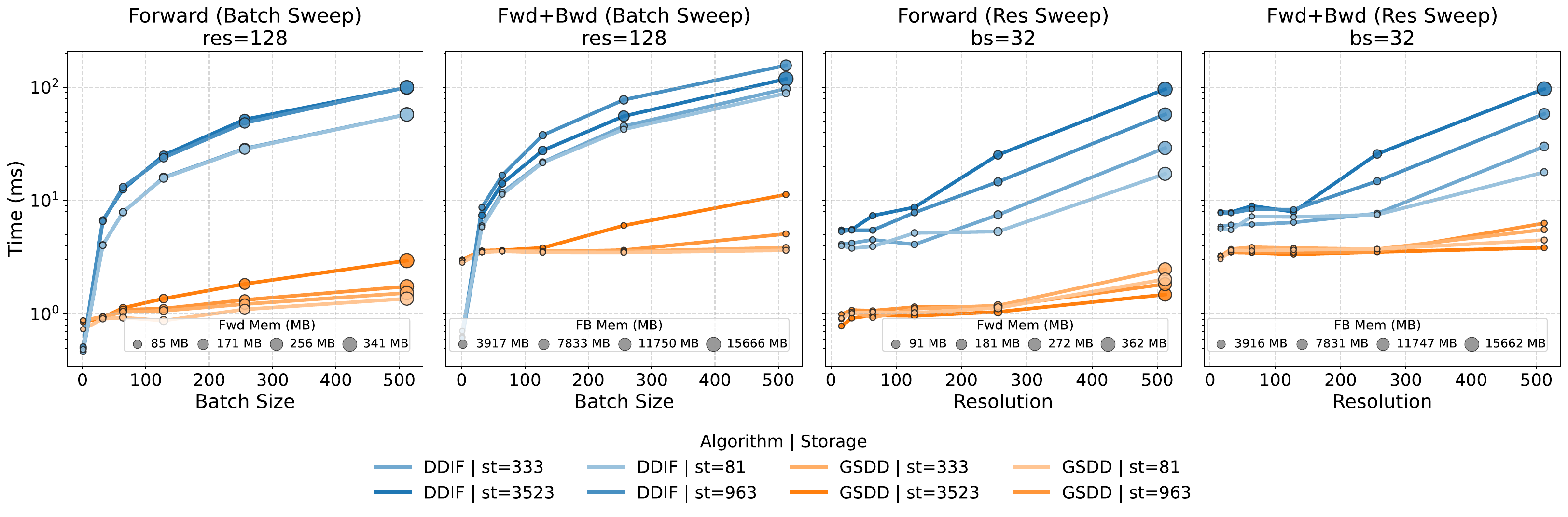}
	\caption{
		\small
	Performance comparison between GSDD and DDiF under matched per-image storage budgets, where st denotes the number of floating-point parameters per image.
	We report forward time, forward plus backward time, and peak GPU memory.
	\textbf{Left}: sweeping image resolution with batch size fixed to 32.
	\textbf{Right}: sweeping batch size with resolution fixed to 128. GSDD achieves consistently lower latency and memory usage, with the advantage becoming more pronounced at higher resolutions and larger batch sizes.
	}
	\label{fig:sweep_all}
\end{figure*}

To comprehensively evaluate the efficiency of the GSDD parameterization, we conduct a quantitative comparison with the state-of-the-art DDiF~\cite{shin2025distilling} in terms of inference and training performance, which corresponds to downstream model training efficiency and dataset distillation efficiency respectively. As illustrated in Figure~\ref{fig:sweep_all}, we benchmark the inference latency and peak GPU memory across varying storage budgets, resolutions, and batch sizes. The results demonstrate that GSDD significantly outperforms DDiF in both decoding speed and memory efficiency. This advantage becomes particularly pronounced when handling high-resolution data or large batch sizes, where GSDD achieves reductions in computational time and memory overhead ranging from several times to an order of magnitude. These findings validate that our method maintains potent representation capabilities while offering exceptional rendering speeds and minimal memory requirements, thereby substantially enhancing the scalability of dataset distillation for practical applications.

\subsection{Results on Cross-Resolution Generalization}
Gaussian Splatting naturally supports rendering at arbitrary resolutions through supersampling, enabling us to distill data at a low resolution (e.g., 128) while still using it to train models at a higher resolution (e.g., 256). This effectively reduces the computation when distilling datasets at large resolutions. Following results and setup in DDiF~\cite{shin2025distilling}, we conduct cross-resolution experiments as shown in Table~\ref{tab:cross_resolution}.

\section{Parallelization Strategy for Batched Rendering}
\label{appendix:parallel}

To scale the rendering pipeline from a single image to an entire batch of images while maximizing GPU throughput, we designed a global ID system coupled with corresponding data structure management. The core idea is to consolidate the rendering of multiple independent images into a single, massive computational task.

\subsection{Core Challenge and the Global ID System}

The parallelization strategy for batched rendering evolves the original single-image pipeline's local indexing system into a global, batch-aware framework. The original approach operated in a local context where screen tiles were indexed from 0 to M-1 for a single image, and data structures pertained only to that instance.

To scale this process across a batch of B images, we introduced a Globally Unique ID (GUID) system. This system creates a single, contiguous address space for all tiles across the entire batch. A local tile j from image i is re-indexed into a global tile id using the linear transformation:
\begin{equation}
	\label{eq:global_id}
	\text{global\_tile\_id} = i \times M + j
\end{equation}
This re-indexing prompted adaptations to our key CUDA kernels. The intersection mapping kernel was modified to calculate this global tile id for each generated intersection record, enabling the subsequent sorting operation to correctly group tasks on a global, batch-wide level. Correspondingly, the rasterization kernel is now launched with a 1D grid of B×M thread blocks, where each block's index directly corresponds to a global tile id. Inside the kernel, this global ID is decomposed back into its constituent image id and local tile id. This allows each thread block to precisely identify which tile of which image it is responsible for, ensuring it writes the final pixel color to the correct memory offset within the batch output tensor.

To support this parallelization strategy, we further employ specific data structures at the Python (PyTorch) level to manage and transfer data efficiently.

\subsection{1D Flattening and Contiguous Memory Layout}
We avoid using Python lists or non-contiguous tensors to store the parameters of different images. Instead, all Gaussian parameters (e.g., means, features, and Cholesky components) across the batch are concatenated and stored in a single large contiguous PyTorch tensor. For a batch of $B$ images with $N$ points each, the means tensor has a shape of $(B \times N, 2)$, rather than being represented as a list of $B$ tensors of shape $(N, 2)$. This design ensures memory contiguity and eliminates the performance overhead caused by frequent concatenation.

By combining the global ID system at the CUDA kernel level with a contiguous memory layout at the framework level, we have constructed an end-to-end rendering architecture that achieves high throughput for large-scale batched rendering tasks.

\subsection{Benchmark across different GPU architectures}
To evaluate the stability of our CUDA-based rasterizer, we conduct systematic benchmarking on five different GPU architectures. We use an image-fitting task in which Gaussian representations reconstruct 1000 images sampled from ImageNet (10 classes $\times$ 100 images in total). Each experiment is repeated three times, and we average all results. We test four storage budgets (21, 170, 682, 2730 Gaussians per image). All experiments used PyTorch 2.5.1 and CUDA $\leq$ 12.0.

\begin{table}[htbp]
  \centering
  \caption{Runtime, memory usage, and PSNR of our rasterizer benchmarked across five GPU architectures under varying storage budgets.}
  \resizebox{\linewidth}{!}{
    \begin{tabular}{l|cccc|cccc|cccc}
    \toprule
          & \multicolumn{4}{c|}{Time (s)} & \multicolumn{4}{c|}{Memory (GB)} & \multicolumn{4}{c}{PSNR} \\
    \multicolumn{1}{l|}{GPU Model} & 21    & 170   & 682   & 2730  & 21    & 170   & 682   & 2730  & 21    & 170   & 682   & 2730  \\
    \midrule
    Tesla-V100-SXM2-32GB & 23.65  & 26.11  & 30.28  & 42.94  & 1.66  & 1.70  & 1.81  & 2.28  & 20.24  & 25.93  & 30.65  & 36.09  \\
    NVIDIA-A100-SXM4-40GB & 10.45  & 12.87  & 19.09  & 41.53  & 1.66  & 1.70  & 1.81  & 2.28  & 20.24  & 26.03  & 31.06  & 36.08  \\
    NVIDIA-GeForce-RTX-3090 & 10.33  & 16.07  & 25.44  & 61.38  & 1.66  & 1.70  & 1.81  & 2.28  & 20.24  & 26.03  & 31.06  & 36.08  \\
    NVIDIA-GeForce-RTX-4090 & 7.60  & 9.38  & 14.06  & 31.00  & 1.66  & 1.70  & 1.81  & 2.28  & 20.24  & 26.02  & 31.03  & 36.06  \\
    NVIDIA-GeForce-RTX-5090 & 7.13  & 6.99  & 9.54  & 21.79  & 1.66  & 1.70  & 1.81  & 2.28  & 20.23  & 26.02  & 31.03  & 36.06  \\
    \bottomrule
    \end{tabular}%
  }
  \label{tab:benchmark_arch}%
\end{table}

As shown in Table~\ref{tab:benchmark_arch}. Different GPUs exhibit different runtimes due to heterogeneous compute capabilities, and RTX5090 is the fastest, while RTX3090 is the slowest. Memory usage remained identical across all platforms, and PSNR variation is small across architectures. Slight performance drops on V100 are expected due to older hardware, but results on A100/3090/4090/5090 remain highly consistent.

\begin{table}[htbp]
  \centering
  \caption{Test accuracies with different resolutions and networks. Images are first distilled with low resolution(128) and tested at higher resolution(256/512).}
	{
    \begin{tabular}{cccccc}
    \toprule
    \textbf{test resolution} & \textbf{test network} & \textbf{method} & \textbf{accuracy↑} & \textbf{difference↓} & \textbf{ratio↓} \\
    \midrule
    \multirow{12}[4]{*}{256} & \multirow{6}[2]{*}{ConvNetD5} & Vanilla & 31.2  & 20.2  & 0.39  \\
          &       & IDC   & 55.0  & 6.4   & 0.10  \\
          &       & SPEED & 58.8  & 8.1   & 0.12  \\
          &       & FreD  & 56.4  & 10.4  & 0.16  \\
          &       & DDiF  & 66.3  & \textbf{5.7 } & \textbf{0.08 } \\
          &       & GSDD  & \textbf{66.4 } & 10.0  & 0.13  \\
\cmidrule{2-6}          & \multirow{6}[2]{*}{ConvNetD6} & Vanilla & 44.0  & 7.3   & 0.14  \\
          &       & IDC   & 55.4  & 6.0   & 0.10  \\
          &       & SPEED & 62.6  & 4.3   & 0.06  \\
          &       & FreD  & 61.8  & 5.0   & 0.07  \\
          &       & DDiF  & 70.6  & \textbf{1.4 } & \textbf{0.02 } \\
          &       & GSDD  & \textbf{71.6 } & 4.8   & 0.06  \\
    \midrule
    \multirow{12}[4]{*}{512} & \multirow{6}[2]{*}{ConvNetD5} & Vanilla & 27.4  & 24.0  & 0.47  \\
          &       & IDC   & 39.5  & 21.9  & 0.36  \\
          &       & SPEED & 45.0  & 21.9  & 0.33  \\
          &       & FreD  & 42.9  & 23.9  & 0.36  \\
          &       & DDiF  & 58.7  & \textbf{13.3 } & \textbf{0.18 } \\
          &       & GSDD  & \textbf{59.4 } & 17.0  & 0.22  \\
\cmidrule{2-6}          & \multirow{6}[2]{*}{ConvNetD6} & Vanilla & 41.2  & 10.1  & 0.20  \\
          &       & IDC   & 51.5  & 9.9   & 0.16  \\
          &       & SPEED & 60.1  & 6.8   & 0.10  \\
          &       & FreD  & 56.3  & 10.5  & 0.16  \\
          &       & DDiF  & \textbf{69.0 } & \textbf{3.0 } & \textbf{0.04 } \\
          &       & GSDD  & 67.8  & 8.6   & 0.11  \\
    \bottomrule
    \end{tabular}%
	}
  \label{tab:cross_resolution}%
\end{table}%

For IDC, SPEED, and FreD, we adopt their best-performing upsampling strategies. For DDiF and GSDD, supersampling is applied, allowing lossless upscaling of the distilled images. GSDD maintains competitive performance in cross-resolution tasks. Although its relative performance drop is larger than DDiF, which is likely due to GSDD's inherent sparsity, reducing its ability to capture fine-grained features when evaluated at higher resolutions.

\section{Visualization of Anti-aliasing and Gaussian Escape}
\label{visualization_antialias_escape}

\begin{figure}[t]
	\centering
	\begin{minipage}[]{0.19\textwidth}
		\centering
		\includegraphics[width=\linewidth]{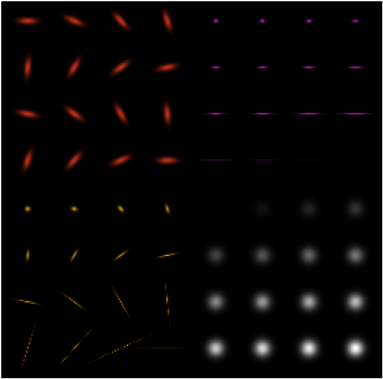}
		
		\includegraphics[width=\linewidth]{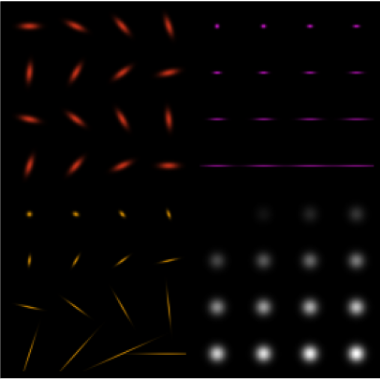}
		\caption*{(a)}
	\end{minipage}%
	\begin{minipage}[]{0.79\textwidth}
		\centering
		\includegraphics[width=0.49\linewidth]{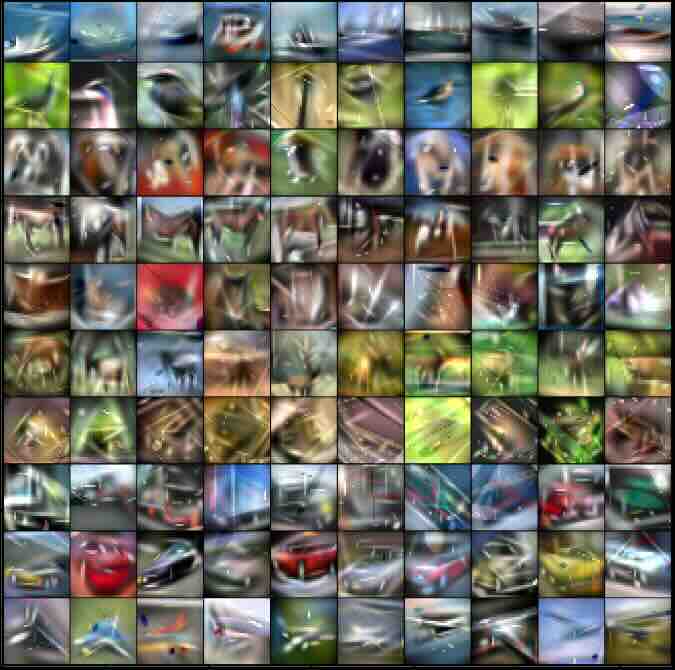}
		\includegraphics[width=0.485\linewidth]{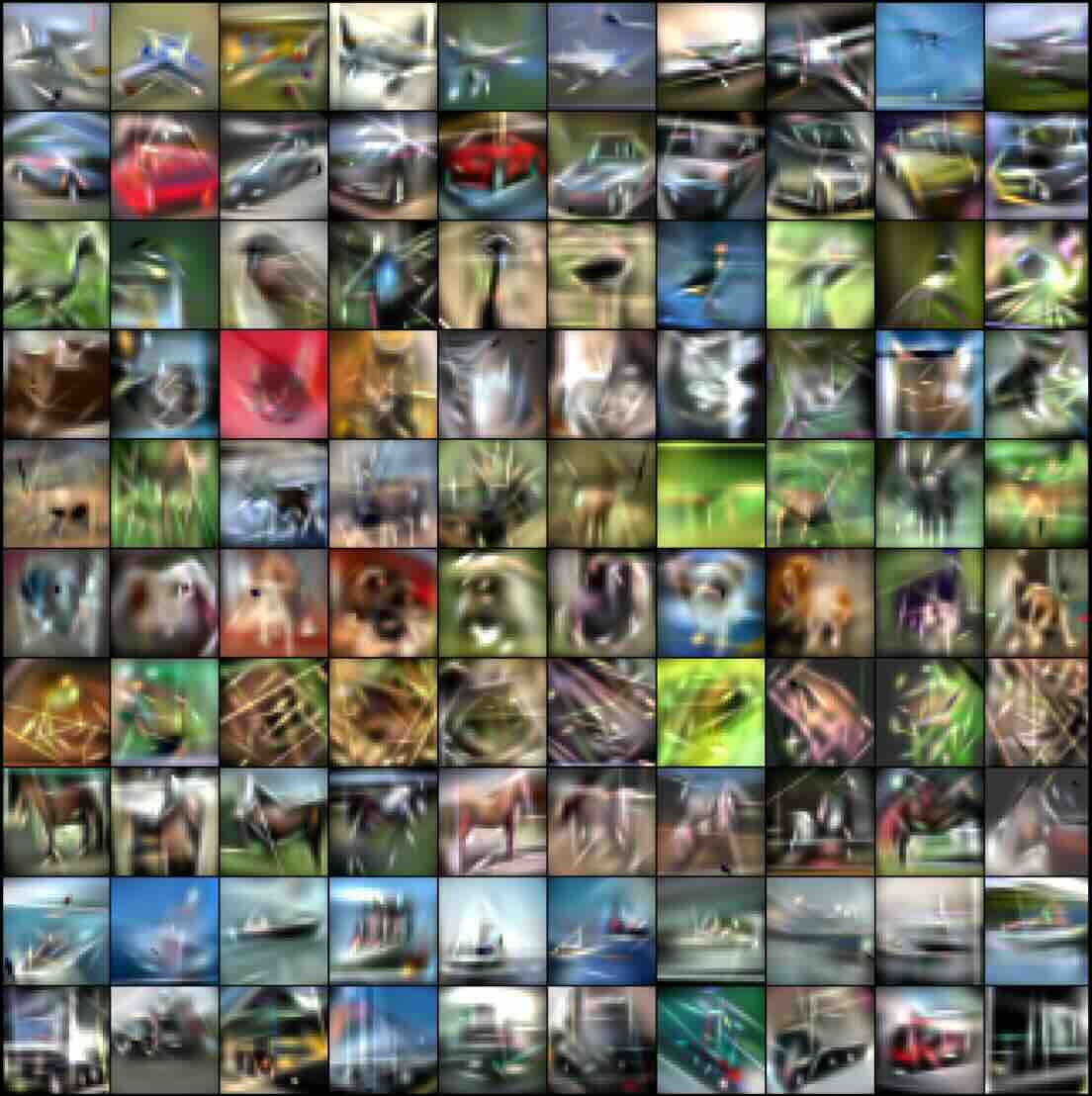}
		\caption*{(b)}
	\end{minipage}
	\caption{Comparison between aliasing and anti-aliasing. 
		(a) shows the schematic illustration, while (b) demonstrates the effect on CIFAR-10 dataset visualization. 
		The anti-aliasing strategy leads to smoother and clearer patterns.}
	\label{fig:aliasing_cifar}
\end{figure}

\begin{figure}[htbp]
	\centering
	\begin{subfigure}[b]{0.48\linewidth}
		\centering
		\includegraphics[width=\linewidth]{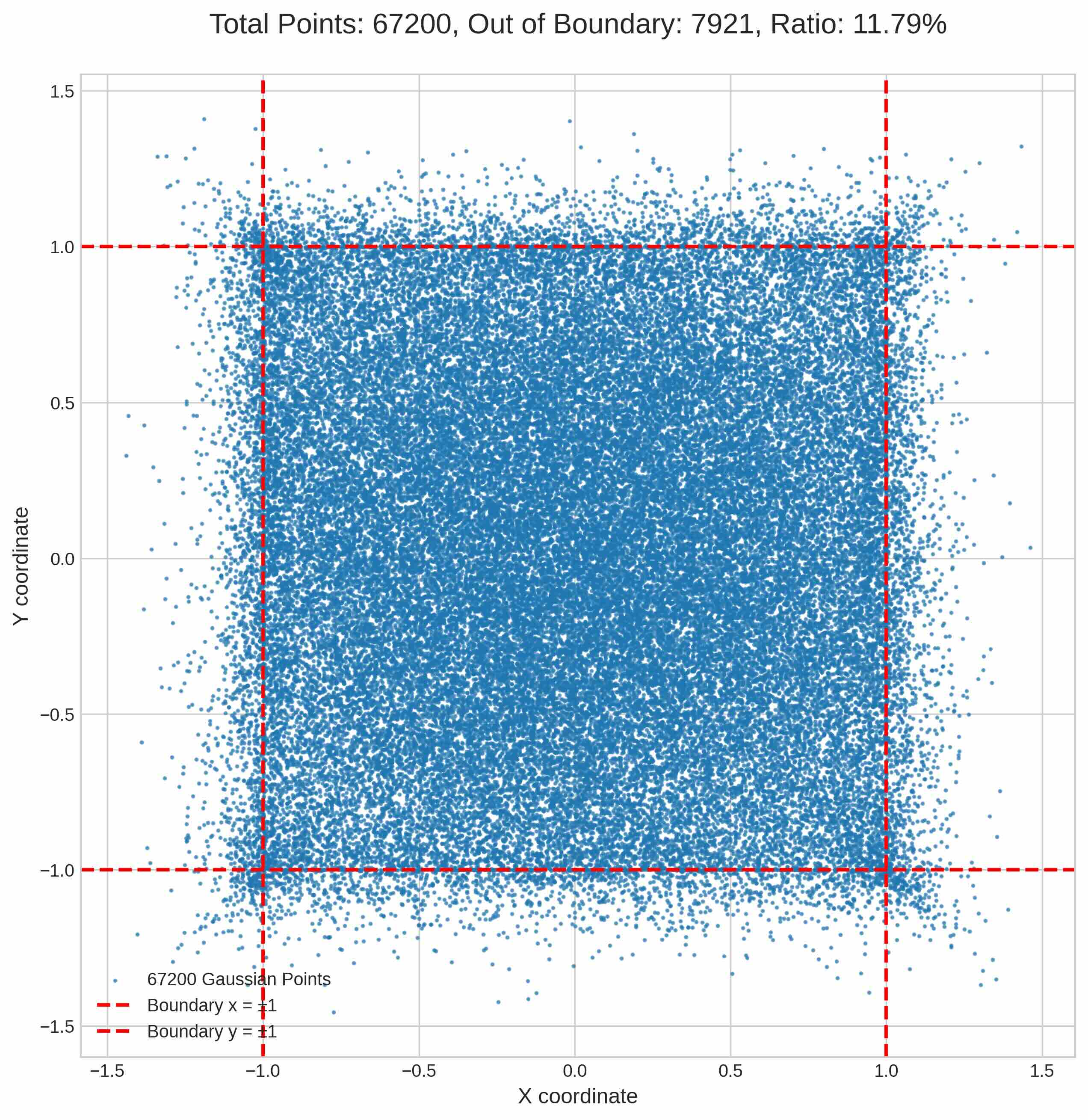}
		\caption{CIFAR-10}
	\end{subfigure}
	\hfill
	\begin{subfigure}[b]{0.48\linewidth}
		\centering
		\includegraphics[width=\linewidth]{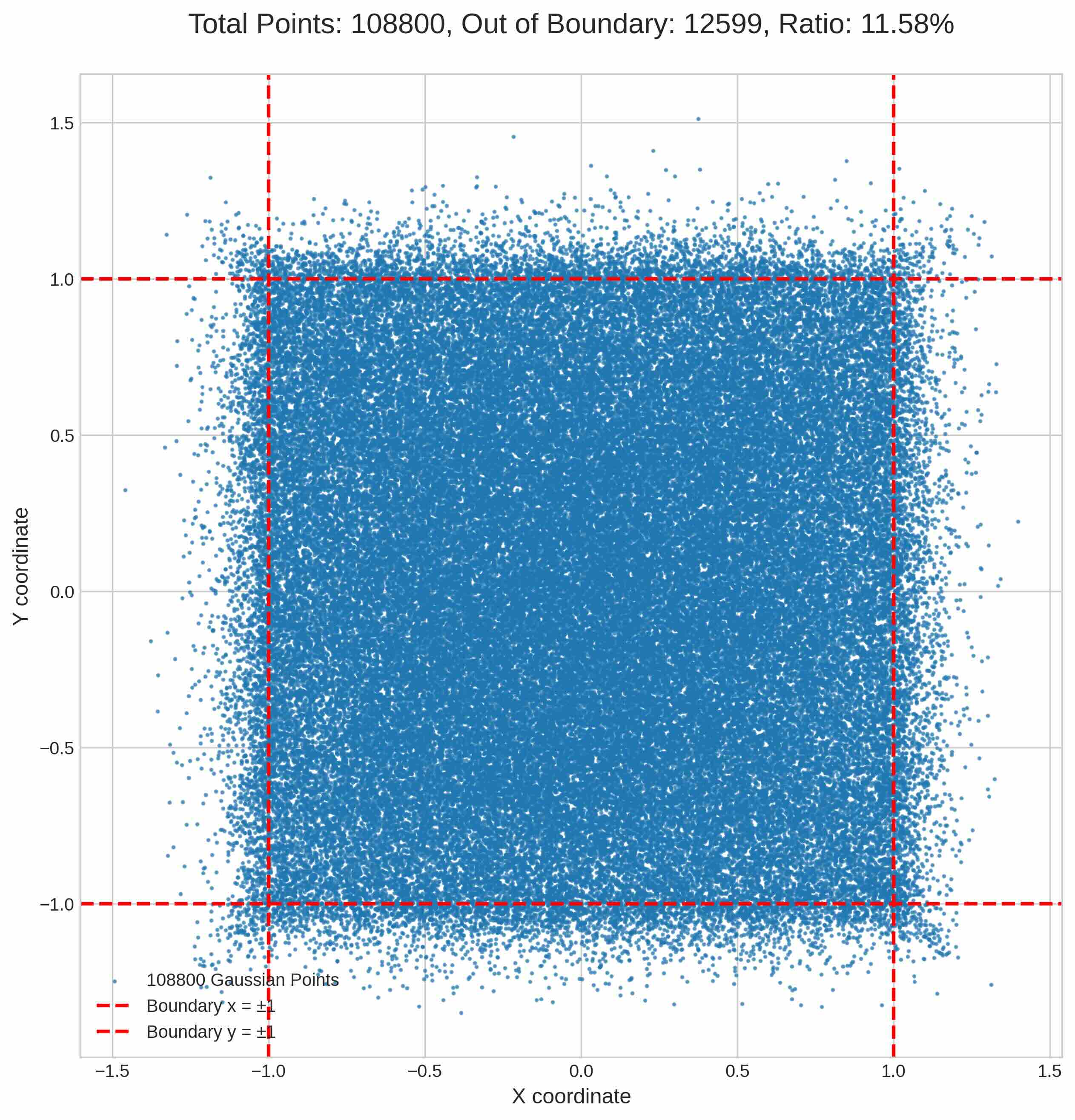}
		\caption{Imagenette}
	\end{subfigure}
	\caption{ 
	Gaussian point centers after optimization without boundary constraints. We plot the mean coordinates of all Gaussian points and observe that many points escape the valid image domain on both CIFAR-10 and ImageNette, which reduces the representational efficiency of the Gaussian budget.
	}
	\label{fig:gaussian_boundary_escape}
\end{figure}


Figure~\ref{fig:aliasing_cifar} visualizes the test images without and with anti-aliasing, along with the actual renderings on CIFAR-10. It is evident that when the Gaussian points are relatively elongated, the rendering quality degrades, resulting in pronounced aliasing artifacts and discontinuous stripes, which in turn adversely affect the performance of the distilled dataset.

In Figure~\ref{fig:gaussian_boundary_escape}, we visualize the escaping phenomenon of Gaussian points after training when no boundary constraints are applied. We plot the distributions of all Gaussian point centers and observe that, on both CIFAR-10 and Imagenette datasets, Gaussian points tend to escape beyond the valid range, which in turn undermines their representational efficiency.

\section{Visualizing GSDD Optimization and Distilled Samples}
\label{visualization_gsdd}

\begin{figure}[htbp]
	\centering
	\includegraphics[width=\textwidth]{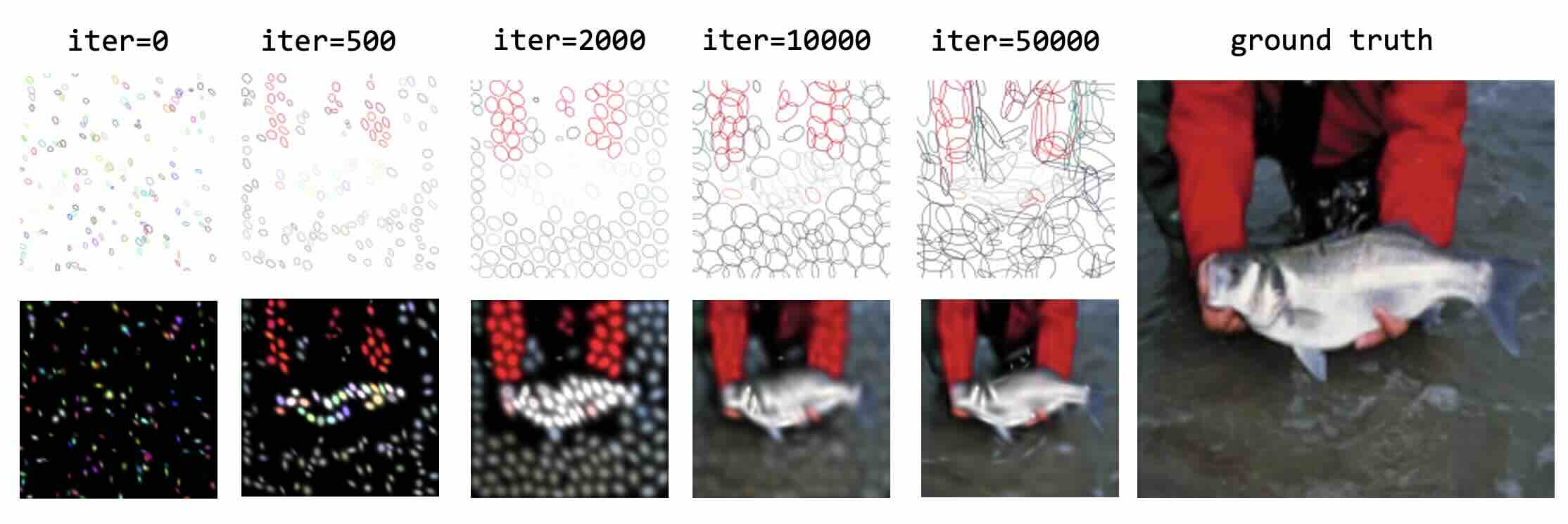}
	\caption{Visualization of the Gaussian initialization process with 136 Gaussian points. 
		The first row shows the actual renderings of Gaussian points at different iterations, 
		while the second row represents each Gaussian as an ellipse encoding its position, shape, and color. 
		During optimization, Gaussian points continuously adjust their positions, shapes, and colors to fit the target image.}
	\label{fig:init_process}
\end{figure}

We visualize the process of Gaussian initialization in Figure~\ref{fig:init_process}.
The first row illustrates the actual rendering results of Gaussian points, while the second row depicts the Gaussians as ellipses that encode their positions, shapes, and colors.
It can be observed that, during optimization, Gaussian points continuously move, reshape, and adjust their colors to fit the target image.

\begin{figure}[htbp]
    \centering
    \begin{subfigure}{0.48\textwidth}
        \centering
        \includegraphics[width=\linewidth]{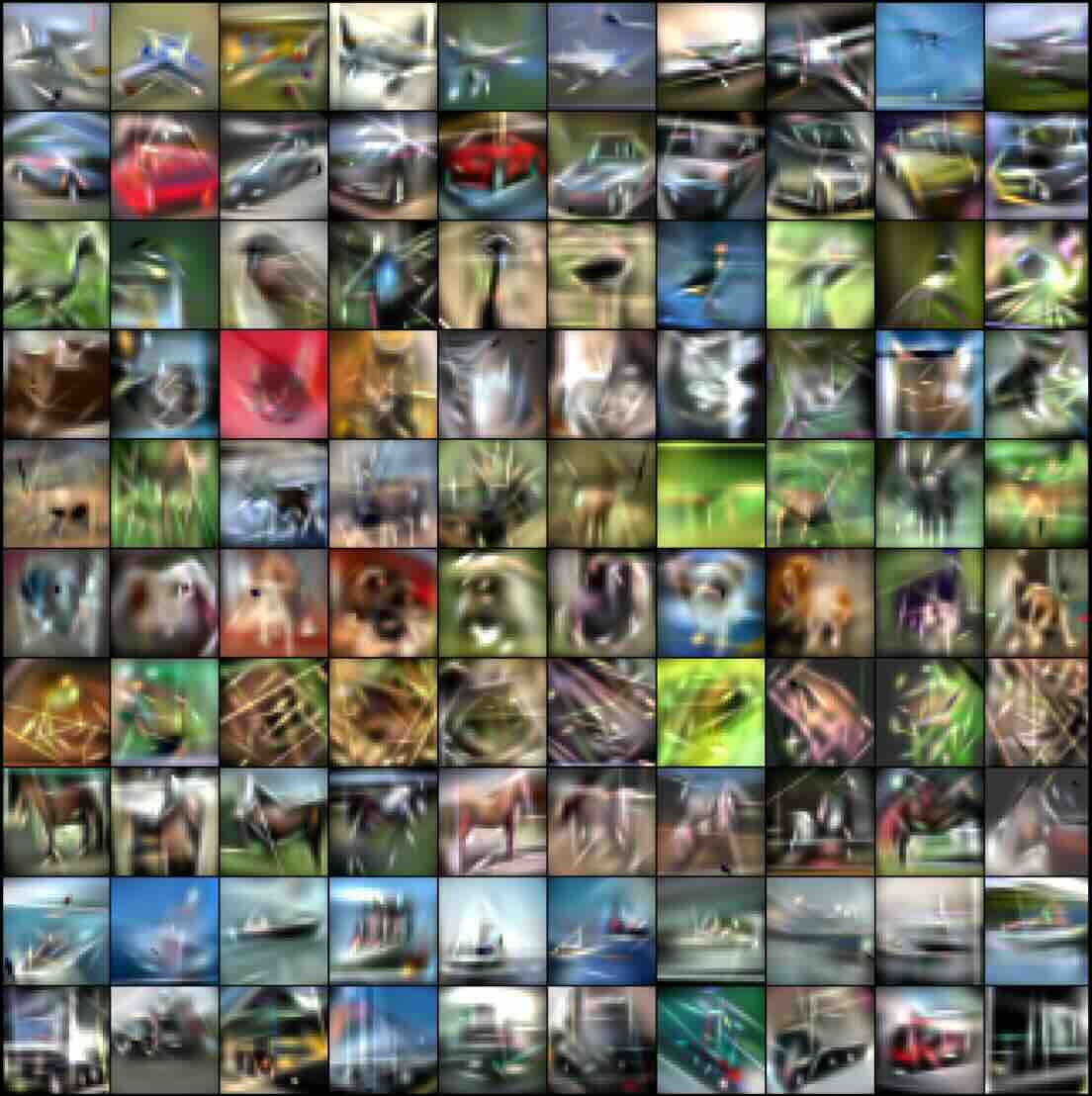}
        \caption{CIFAR-10}
        \label{fig:cifar10}
    \end{subfigure}
    \hfill
    \begin{subfigure}{0.48\textwidth}
        \centering
        \includegraphics[width=\linewidth]{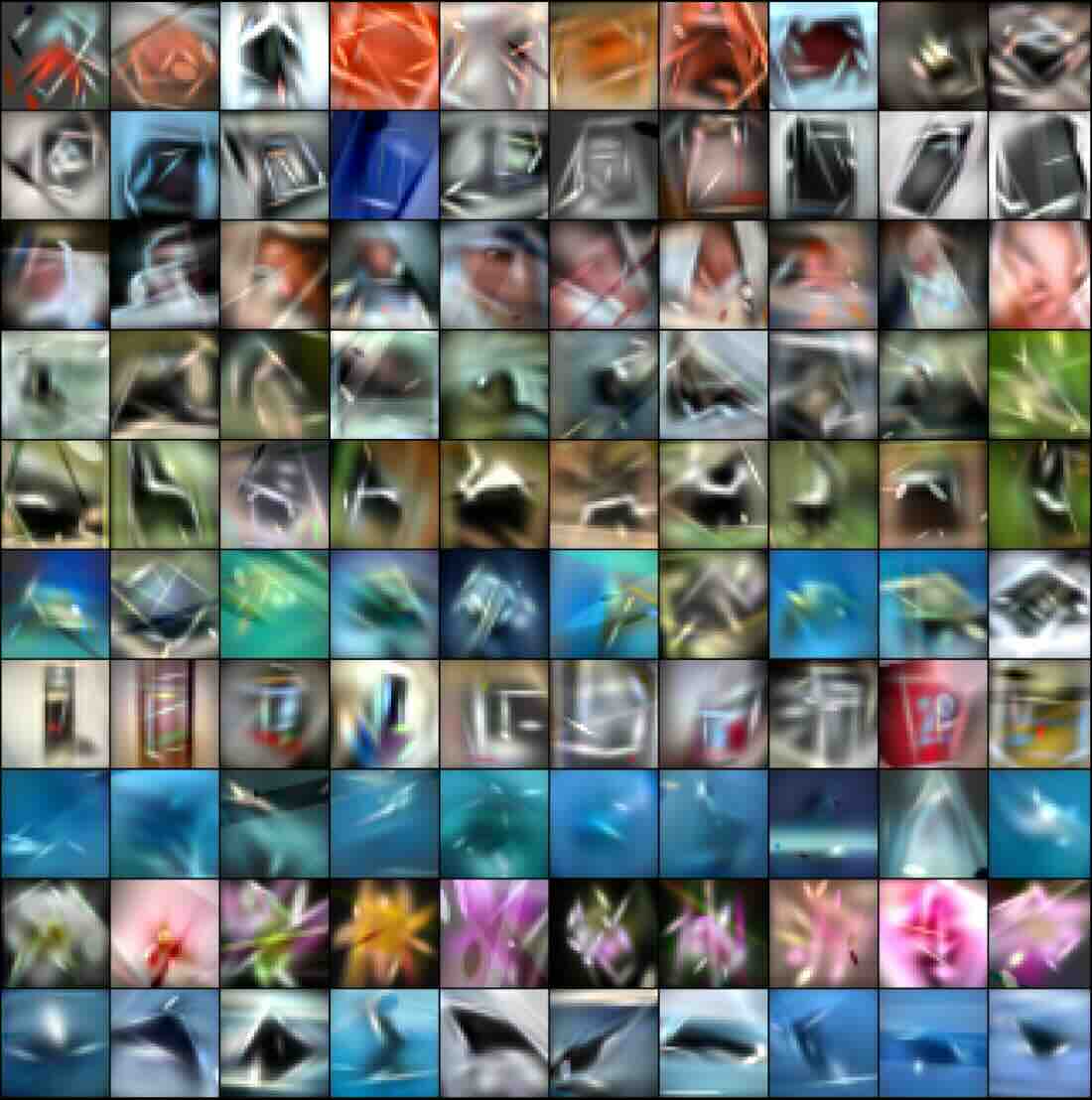}
        \caption{CIFAR-100}
        \label{fig:cifar100}
    \end{subfigure}
    \caption{Visualizations of distilled images produced by TM on CIFAR-10 and CIFAR-100.}
\end{figure}

\begin{figure}[htbp]
    \centering
    \begin{subfigure}{0.48\textwidth}
        \centering
        \includegraphics[width=\linewidth]{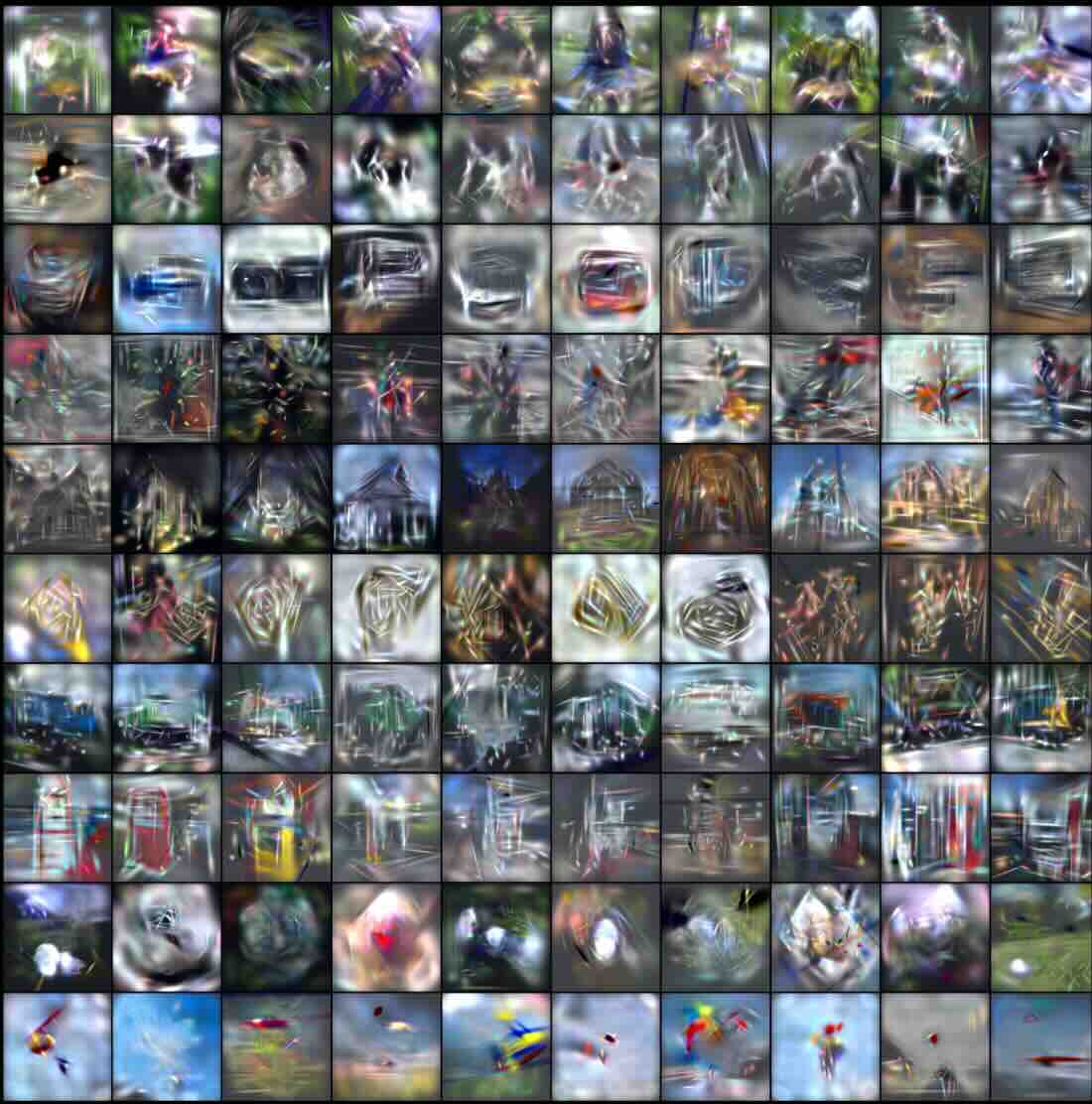}
        \caption{ImageNette (TM)}
        \label{fig:imagenette}
    \end{subfigure}
    \hfill
    \begin{subfigure}{0.48\textwidth}
        \centering
        \includegraphics[width=\linewidth]{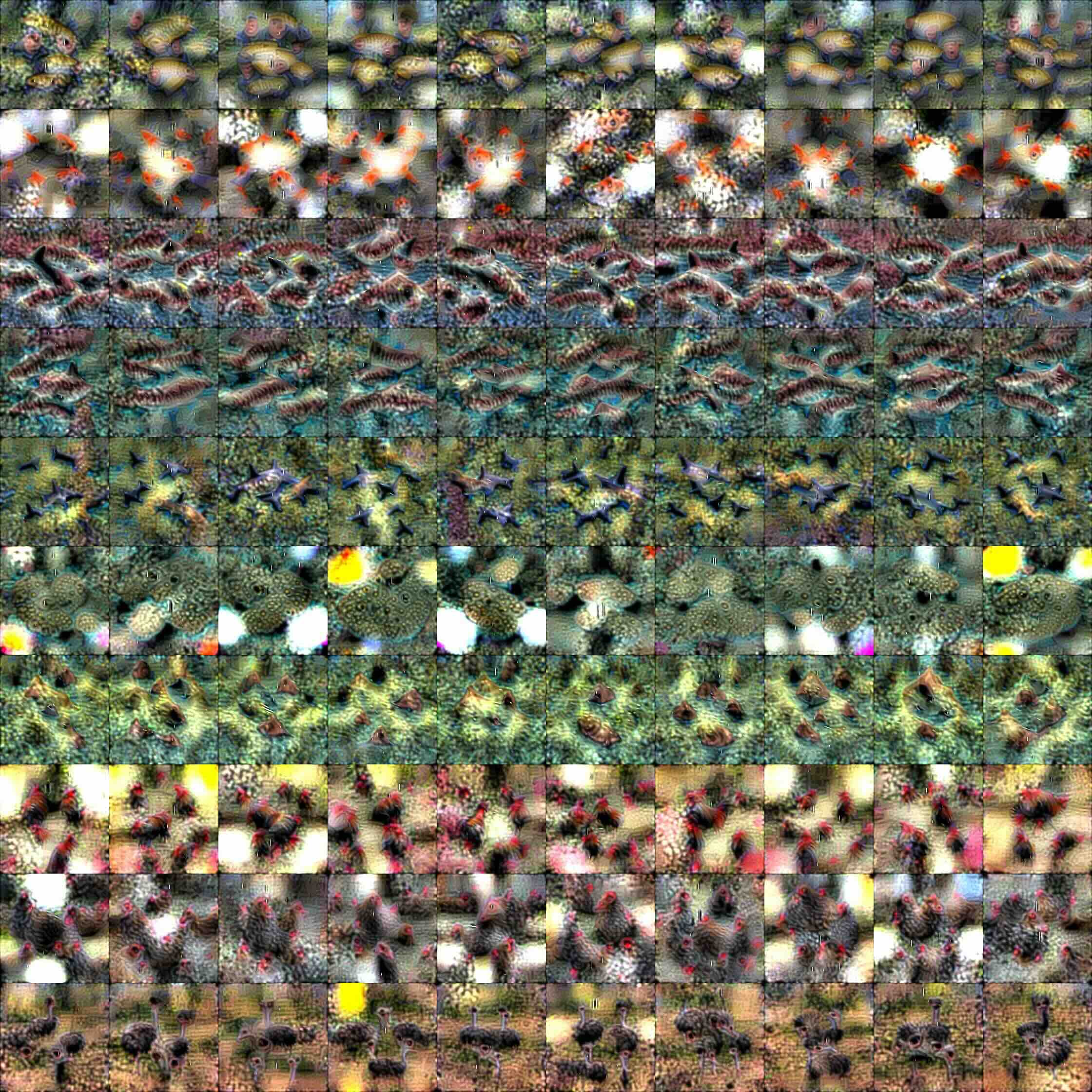}
        \caption{ImageNet-1K (SRe2L)}
        \label{fig:vis_imagenet}
    \end{subfigure}
    \caption{Visualizations of distilled images on ImageNette and ImageNet-1K generated by different distillation methods.}
\end{figure}

\begin{figure}[htbp]
    \centering
    \begin{subfigure}{0.48\textwidth}
        \centering
        \includegraphics[
			width=\linewidth,
			trim=400 150 400 150,
			clip
			]{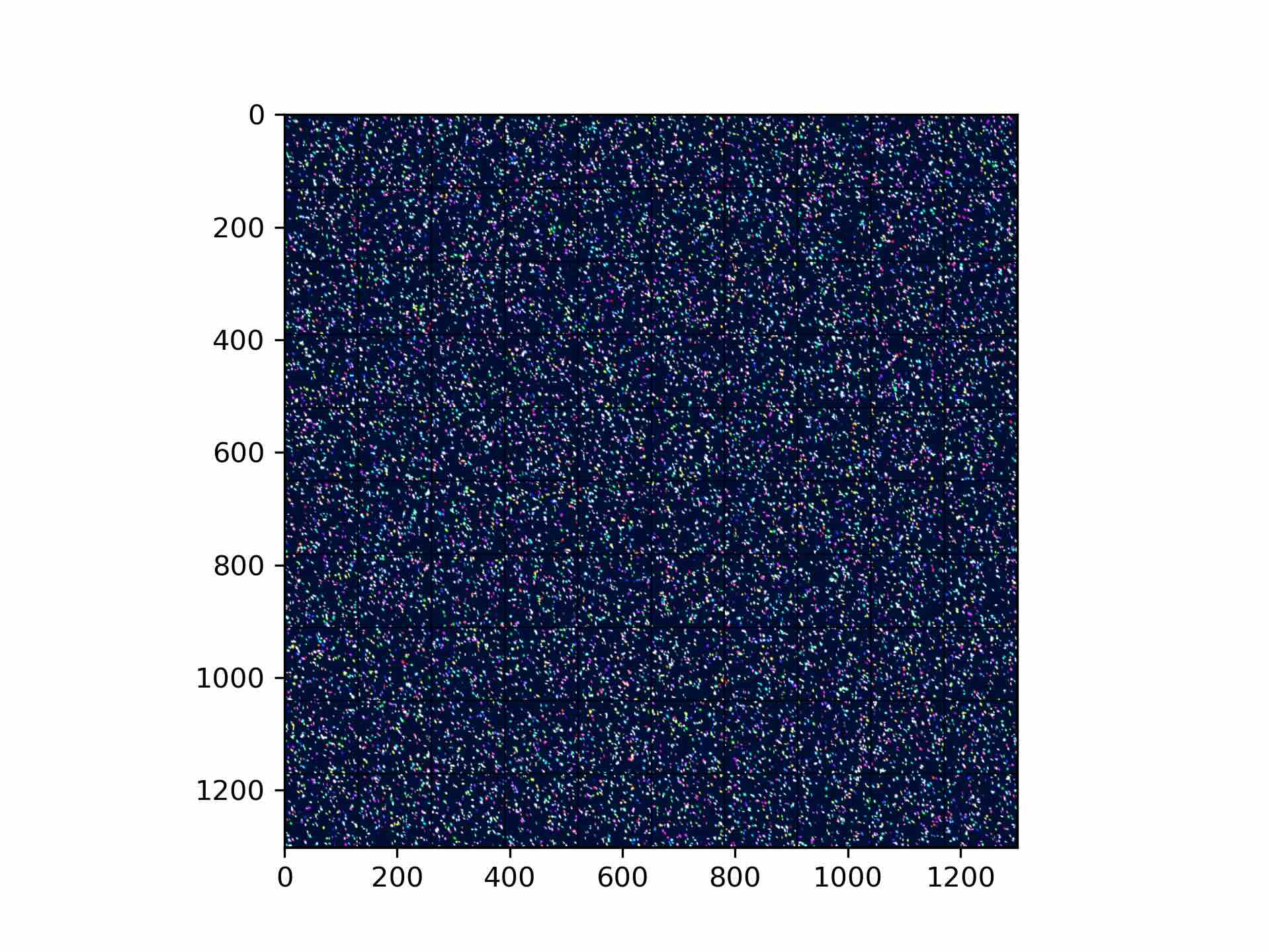}
        \caption{Before distillation}
        \label{fig:vis_random_init}
    \end{subfigure}
    \hfill
    \begin{subfigure}{0.48\textwidth}
        \centering
        \includegraphics[
			width=\linewidth,
			trim=400 150 400 150,
			clip
			]{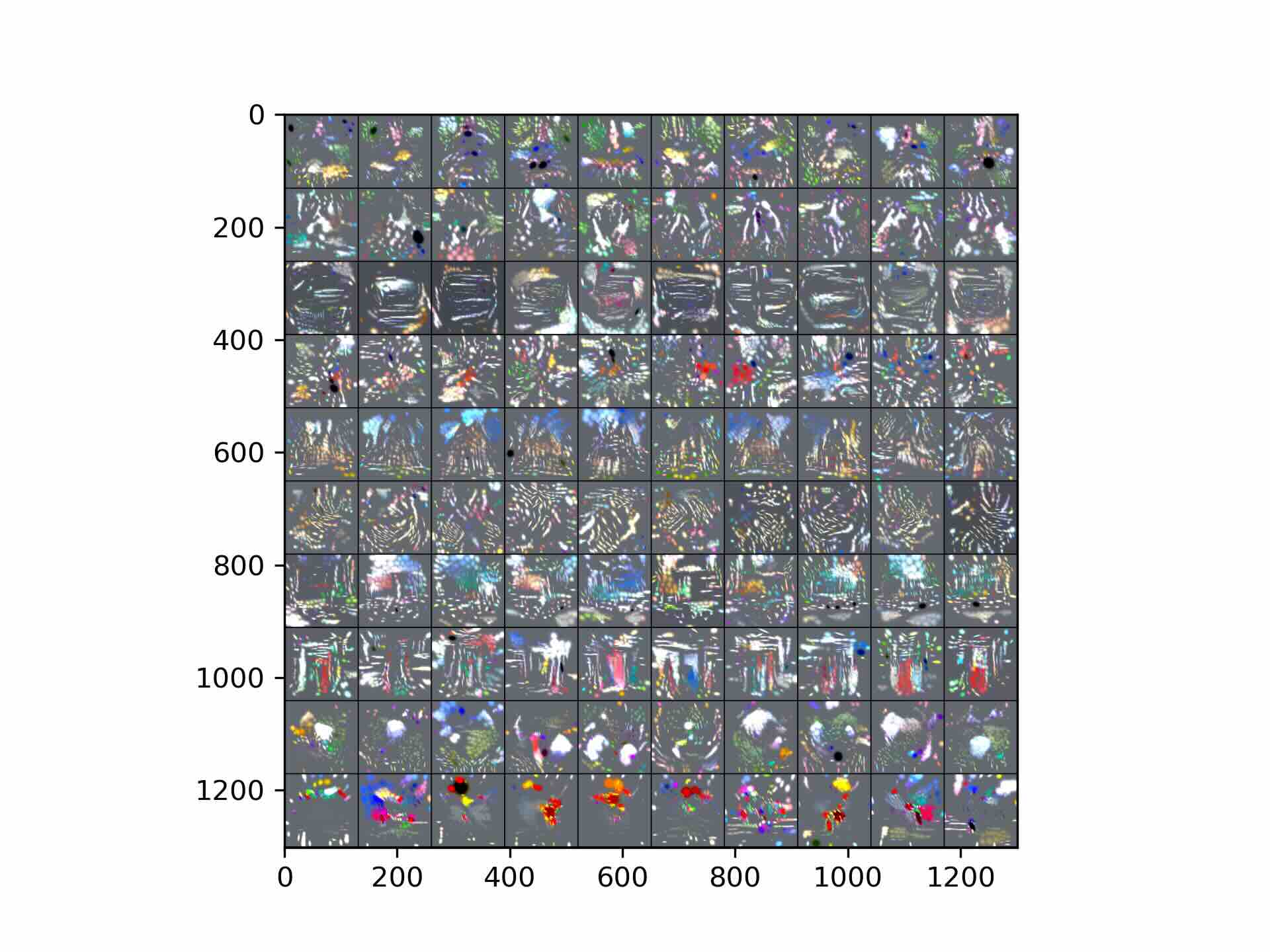}
        \caption{After distillation}
        \label{fig:vis_random_init_final}
    \end{subfigure}
    \caption{Evolution of randomly initialized distilled images before and after the distillation process.}
\end{figure}

\begin{figure}[htbp]
    \centering
    \begin{subfigure}{0.48\textwidth}
        \centering
        \includegraphics[
			width=\linewidth,
			trim=400 150 400 150,
			clip
		]{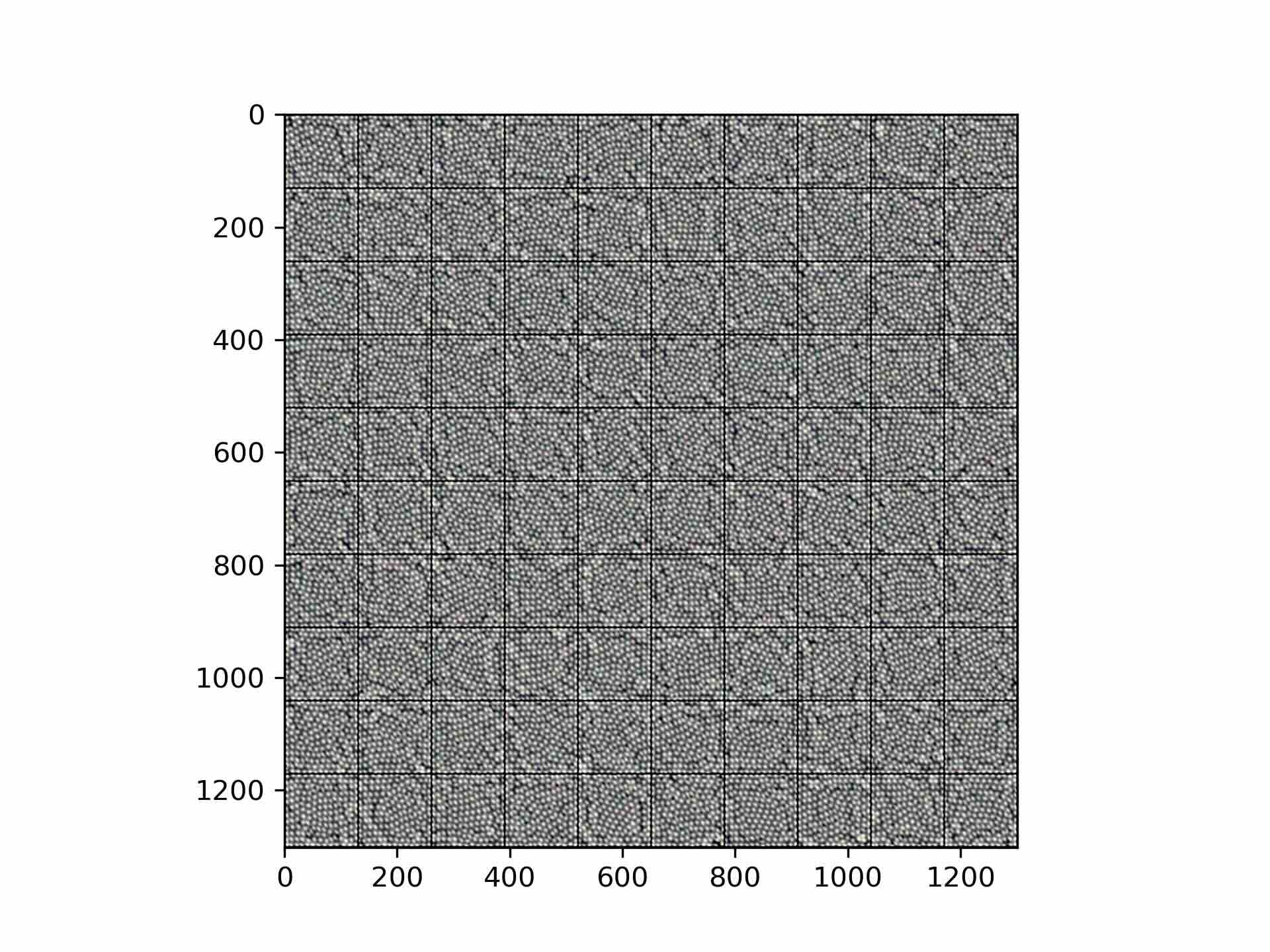}
        \caption{Before distillation}
        \label{fig:vis_warmup_init}
    \end{subfigure}
    \hfill
    \begin{subfigure}{0.48\textwidth}
        \centering
        \includegraphics[
			width=\linewidth,
			trim=400 150 400 150,
			clip
		]{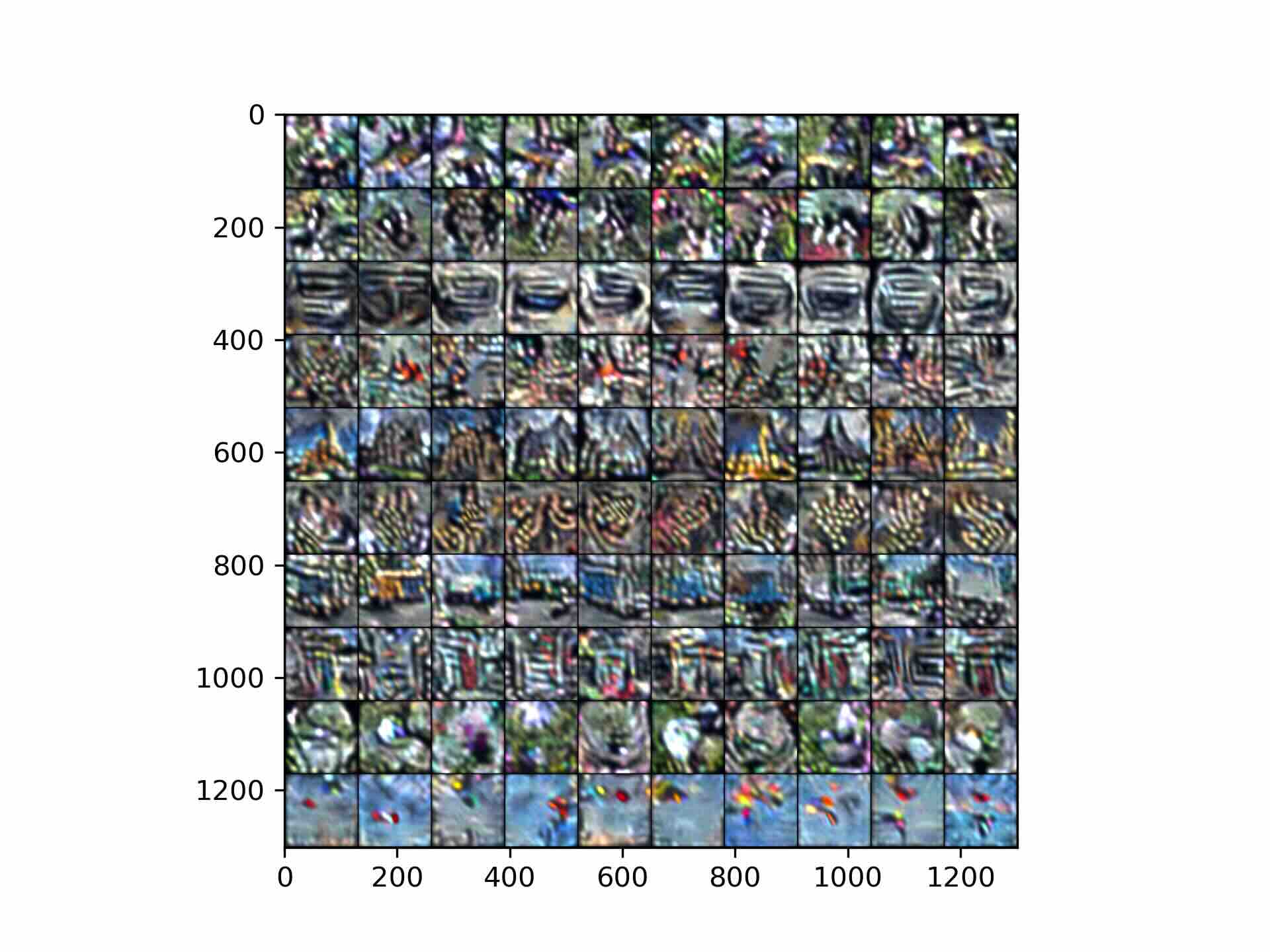}
        \caption{After distillation}
        \label{fig:vis_warmup_init_final}
    \end{subfigure}
    \caption{Evolution of warmup-initialized distilled images throughout the distillation process.}
\end{figure}

\end{document}